\documentclass{article}
\usepackage{url}
\usepackage{soul}
\usepackage[small]{caption}
\usepackage{graphicx}
\usepackage{amsmath}
\usepackage{pifont}
\usepackage{amsthm}
\usepackage{booktabs}
\usepackage{enumitem}
\usepackage{algorithm}
\usepackage{algorithmic}
\usepackage{epstopdf}
\usepackage{makecell}
\usepackage[accsupp]{axessibility}
\usepackage{alltt}
\usepackage{multirow}
\usepackage{wrapfig}
\usepackage{adjustbox}
\usepackage{bbding} 
\usepackage[dvipsnames]{colortbl}
\usepackage{subcaption}
\PassOptionsToPackage{dvipsnames}{xcolor}
\definecolor{olive}{rgb}{0.5, 0.5, 0.0}
\definecolor{maroon}{rgb}{0.69, 0.19, 0.38}
\definecolor{celestialblue}{rgb}{0.29, 0.59, 0.82}
\definecolor{darkgreen}{rgb}{0.0, 0.6, 0.0}
\definecolor{grey}{rgb}{0.5,0.5,0.5}
\definecolor{darkblue}{rgb}{0.19, 0.19, 0.62}
\definecolor{silver}{rgb}{0.7,0.7,0.7}
\definecolor{darkcyan}{rgb}{0.0, 0.55, 0.55}

\def\clap#1{\hbox to 0pt{\hss #1\hss}}%

\newcommand\undefcolumntype[1]{\expandafter\let\csname NC@find@#1\endcsname\relax}

\definecolor{C0}{rgb}{0.121569, 0.466667, 0.705882}
\definecolor{C1}{rgb}{1.000000, 0.498039, 0.054902}
\definecolor{C2}{rgb}{0.172549, 0.627451, 0.172549}
\definecolor{C3}{rgb}{0.839216, 0.152941, 0.156863}
\definecolor{C4}{rgb}{0.580392, 0.403922, 0.741176}
\definecolor{C5}{rgb}{0.549020, 0.337255, 0.294118}
\definecolor{C6}{rgb}{0.890196, 0.466667, 0.760784}
\definecolor{C7}{rgb}{0.498039, 0.498039, 0.498039}
\definecolor{C8}{rgb}{0.737255, 0.741176, 0.133333}
\definecolor{C9}{rgb}{0.090196, 0.745098, 0.811765}

\newcommand{\rr}[1]{\textcolor{black}{#1}}
\newcommand{\bb}[1]{\textcolor{C9}{#1}}

\newcommand{\baseline}[1]{\textit{\textcolor{gray}{#1}}}
 
\newlength\savewidth

\usepackage{amsmath,amsfonts,bm}

\def\eqref#1{equation~\ref{#1}}

\def\1{\bm{1}}

\DeclareMathAlphabet{\mathsfit}{\encodingdefault}{\sfdefault}{m}{sl}
\SetMathAlphabet{\mathsfit}{bold}{\encodingdefault}{\sfdefault}{bx}{n}

\newcommand{\KL}{D_{\mathrm{KL}}}

\definecolor{Blue}{RGB}{0, 0, 255}

\definecolor{Aquamarine}{RGB}{127, 255, 212}
\definecolor{Sepia}{RGB}{112, 66, 20}
\definecolor{BrickRed}{RGB}{203, 65, 84}
\newcommand{\cmark}{\ding{51}}%
\newcommand{\xmark}{\ding{55}}%
\usepackage{microtype}
\usepackage{tikz}
\usepackage{epsfig}
\usepackage{microtype}
\usepackage{mathrsfs}
\usepackage{tabularray}
\usepackage{hyperref}

\usepackage[accepted]{icml2023}

\usepackage{autobreak}
\allowdisplaybreaks
\usepackage{amssymb}
\usepackage{mathtools}

\usepackage[capitalize,noabbrev]{cleveref}

\definecolor{Blue}{RGB}{0, 0, 255}
\definecolor{Aquamarine}{RGB}{127, 255, 212}
\definecolor{Sepia}{RGB}{112, 66, 20}
\definecolor{BrickRed}{RGB}{203, 65, 84}
\colorlet{my-red}{BrickRed!90!Sepia}
\colorlet{my-blue}{Aquamarine!30!Blue}
\hypersetup{
  colorlinks,
  urlcolor  = my-red,
  linkcolor = my-blue,
  citecolor = my-blue,
}
\definecolor{C3}{rgb}{0.839216, 0.152941, 0.156863}

\usepackage{bm}
\usepackage{tabu}

\usepackage[capitalize]{cleveref}
\crefname{section}{Sec.}{Secs.}
\Crefname{section}{Section}{Sections}
\Crefname{table}{Table}{Tables}
\crefname{table}{Table}{Tables}

\theoremstyle{plain}
\newtheorem{theorem}{Theorem}[section]
\newtheorem{proposition}[theorem]{Proposition}

\theoremstyle{definition}

\theoremstyle{remark}

\usepackage[capitalize]{cleveref}
\crefname{section}{Sec.}{Secs.}
\Crefname{section}{Section}{Sections}
\Crefname{table}{Table}{Tables}
\crefname{table}{Table}{Tables}

\usepackage[textsize=tiny]{todonotes}

\icmltitlerunning{Precise Knowledge Transfer via Flow Matching}

\begin{document}

\twocolumn[
\icmltitle{Precise Knowledge Transfer via Flow Matching}



\icmlsetsymbol{equal}{*}

\begin{icmlauthorlist}
\icmlauthor{Shitong Shao}{mbzuai,shai}
\icmlauthor{Zhiqiang Shen}{mbzuai}
\icmlauthor{Linrui Gong}{sjtu}
\icmlauthor{Huanran Chen}{thu}
\icmlauthor{Xu Dai}{shai}
\end{icmlauthorlist}

\icmlaffiliation{mbzuai}{Mohamed bin Zayed University of Artificial Intelligence}
\icmlaffiliation{shai}{Shanghai AI Laboratory}
\icmlaffiliation{thu}{Tsinghua University}
\icmlaffiliation{sjtu}{Shanghai Jiaotong University}

\icmlcorrespondingauthor{Zhiqiang Shen}{Zhiqiang.Shen@mbzuai.ac.ae}

\icmlkeywords{Machine Learning, ICML}

\vskip 0.3in
]



\printAffiliationsAndNotice{}  

\begin{abstract}
In this paper, we propose a novel knowledge transfer framework that introduces continuous normalizing flows for progressive knowledge transformation and leverages multi-step sampling strategies to achieve precision knowledge transfer. We name this framework Knowledge Transfer with Flow Matching (FM-KT), which can be integrated with a metric-based distillation method with any form (\textit{e.g.} vanilla KD, DKD, PKD and DIST) and a meta-encoder with any available architecture (\textit{e.g.} CNN, MLP and Transformer). By introducing stochastic interpolants, FM-KD is readily amenable to arbitrary noise schedules (\textit{e.g.}, VP-ODE, VE-ODE, Rectified flow) for normalized flow path estimation. We theoretically demonstrate that the training objective of FM-KT is equivalent to minimizing the upper bound of the teacher feature map or logit negative log-likelihood. Besides, FM-KT can be viewed as a unique implicit ensemble method that leads to performance gains. By slightly modifying the FM-KT framework, FM-KT can also be transformed into an online distillation framework OFM-KT with desirable performance gains. Through extensive experiments on CIFAR-100, ImageNet-1k, and MS-COCO datasets, we empirically validate the scalability and state-of-the-art performance of our proposed methods among relevant comparison approaches.
\end{abstract}

\vspace{-5pt}
\section{Introduction}
\label{sec:intro}
Despite the remarkable achievements of deep neural networks, the dramatic increase in the number of parameters in recent years prevents their application to real-world scenarios. To solve this problem, knowledge distillation~\citep{vanillakd} has been introduced for model compression in order to deploy lightweight models with desirable performance on mobile devices. One critical component of knowledge distillation is knowledge transfer, which aims to transfer knowledge from the teacher to the student, ensuring efficient student performance during runtime. The vast majority of existing research focuses on enhancing the capability of knowledge transfer, including how to design effective and efficient meta-encoders to transform the output (\textit{i.e.} feature or logit) of the student in the high-dimensional space to match the corresponding output of the teacher~\citep{HSAKD,cvpr22_kd_guided,GSD}, and designing metric-based distillation methods to reduce the gap between the knowledge of the teacher and the knowledge of the student~\citep{DIST,DKD}.

\begin{figure}[t]
\includegraphics[width=0.95\linewidth,trim={0cm 0.2cm 0cm 0cm},clip]{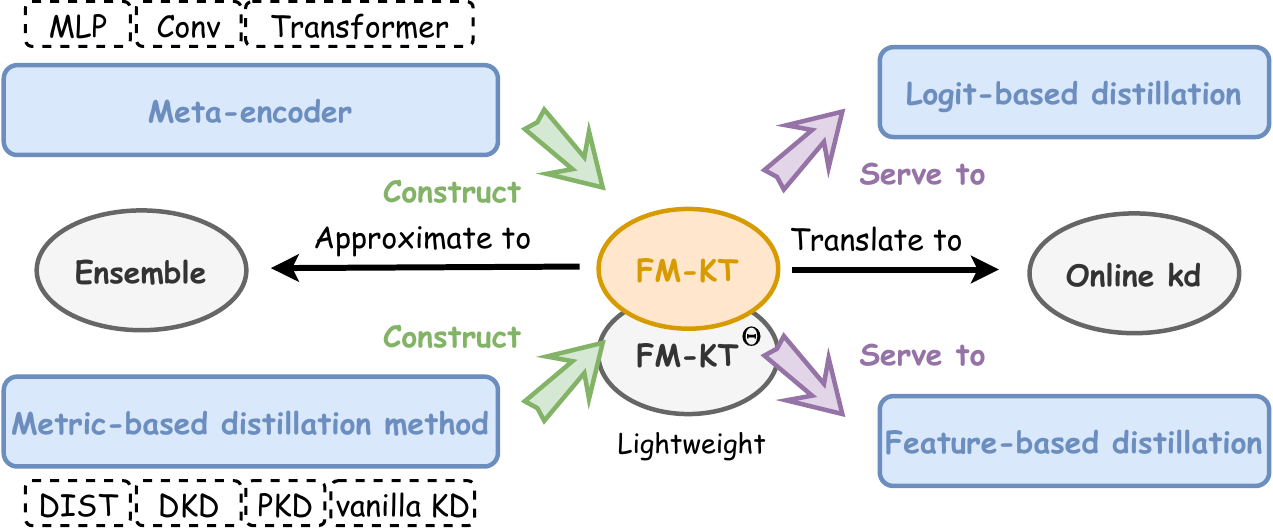}
\caption{A highly scalable knowledge transfer framework FM-KT.}
\label{fig:summarized_of_fm_kd}
\vspace{-10pt}
\end{figure}

\noindent{\bf Motivation.} The significant discrepancy in feature/logit distributions between the teacher and the student adversely impacts distillation performance, a challenge not fully mitigated by even well-designed meta-encoders and metric-based distillation methods~\citep{kdsurvey}. This difficulty stems from the inherent challenges of transferring complex features or logits between the teacher and the student in a single step, which often compromises reliability and precision. A promising strategy involves segmenting the distribution gap into multiple sub-parts and sequentially matching these distributions to facilitate gradual and accurate knowledge transfer~\citep{DiffKD,kdiffusion}. Thus, a question worth exploring is, ``{\textit{How to implement multi-step sampling to facilitate a progressive transformation, thereby achieving more effective and precise knowledge transfer?}}''

In this work, we take apart the features/logits of the teacher and the student as empirical distributions in an attempt to answer this question. From this perspective, diffusion models~\cite{ddpm_begin,sde} and continuous normalized flows (CNFs)~\citep{CNF,stochastic_interpolant} are suitable for implementing progressive transformation. Among them, diffusion models were employed by the prior study DiffKD~\citep{DiffKD} to transition from the student feature/logit distribution to the teacher counterpart. However, the inherent characteristics of diffusion models necessitate that one end of the distribution trajectory adheres to a Gaussian distribution. Consequently, this process requires transforming the student feature/logit distribution into a Gaussian approximation before it can be converted into the teacher feature/logit distribution during reverse sampling. Moreover, the complexity of this approach prevents the full exploitation of its potential in progressive transformation. In contrast, CNFs can directly map these two empirical distributions without accessing uncorrelated Gaussian distributions. This allows for the progressive transformation to reshape the student's feature/logit density to the teacher's with more fine-grained information.

Unfortunately, directly implementing this approach faces a critical challenge: popular flow matching~\cite{iclr22_rect,CNF,stochastic_interpolant}, akin to the evidence lower bound (ELBO) in diffusion models~\citep{ddpm_begin,ddim,sde}, tends to an unreliable alignment with a high probability space of the teacher's knowledge. This approach inadvertently informs the student and meta-encoder about the target information the teacher's knowledge during the training phase, which can be considered as ``cheating''. This approach substantially undermines the generalization ability of the student.

To address this issue and realize precise knowledge transfer, we propose a novel framework, Knowledge Transfer with Flow Matching (FM-KT), to amend incorrect single output from the student through multi-step sampling. To be specific, we design a serial training paradigm with theoretical guarantees to avoid ``cheating'' in knowledge distillation, finally yielding a reliable meta-encoder for multi-step sampling during inference. It is worth mentioning that by changing the noise schedule~\citep{kingma2021variational} in CNFs, FM-KT can consistently model various probabilistic flows, such as VP ODE~\citep{iclr22_rect,sde}, VE ODE~\citep{iclr22_rect,sde}, and Rectified flow~\citep{iclr22_rect,CNF}.

FM-KT is a versatile training paradigm for knowledge transfer with high scalability. As depicted in Fig.~\ref{fig:summarized_of_fm_kd}, FM-KT is comprised of a meta-encoder with any available architecture and a metric-based distillation method with any form, enabling both feature-based and logit-based distillation, and consequently enhancing the generalization ability of the student. Most importantly, it can be theoretically interpreted as an implicit ensemble algorithm when the noise schedule is set as Rectified flow. Notably, we propose a variant of FM-KT called FM-KT$^{\Theta}$, which avoids additional computational overhead during inference. By introducing a metric function between the predicted velocity at each time point and the numerical solution derived from the final discrete sampling, FM-KT can be transformed into an online distillation algorithm OFM-KT. Our experiments, both qualitative and quantitative, demonstrate that our proposed methods enhance performance across both image classification and object detection tasks.

\vspace{-5pt}
\section{Preliminaries}
\label{sec:pre}

\paragraph{Review the Knowledge Transfer.} Knowledge transfer plays an important role in knowledge distillation~\citep{vanillakd}, which aims to transfer the teacher's knowledge to the student, thus enhancing the performance of the student. \rr{In classical knowledge distillation algorithms, a common and simple approach~\citep{ATKD,VID,SPKD,SSKD,CRD,PKD,DKD,DIST} is to align the student's feature/logit $X^S$ with the teacher's feature/logit $X^T$ using two encoders: $g^S(\cdot)$ and $g^T(\cdot)$.} This can be expressed as $\min L(g^S(X^S),g^T(X^T))$, where $L(\cdot,\cdot)$ refers to the distance metric function \rr{with any form}. \rr{In some cases}, $g^S(\cdot)$ and $g^T(\cdot)$ can be reduced to \textit{the identity function}, making the supervision a direct matching between $X^S$ and $X^T$. This is widely employed in logit-based distillation.

\vspace{-3pt}
Recently, DiffKD and KDiffusion were introduced by~\citet{DiffKD} and~\citet{kdiffusion}, which employ a diffusion model (\textit{i.e.} meta-encoder) $g_{v_\theta}(\cdot)$ to replace the original $g^S(\cdot)$ for more effective progressive transformation, which significantly enhances the generalization ability of the student. However, DiffKD does not immediately effectuate the transition from $X^S$ to $X^T$. Instead, it first transforms $X^S$ into Gaussian noises and subsequently translates this noise to $X^T$. This dual-stage transformation process might potentially be overly complex, substantially increasing the inference cost and hindering its widespread use. KDiffusion complicates its noise schedule, and the supervision of task information is sparse (\textit{i.e.} $\mathcal{L}_\textrm{guided}$ in the original paper), leading to its relatively poor performance in ImageNet-1k.

\vspace{-10pt}
\paragraph{Continuous Normalized Flows.} Given the couple $(X^S,X^T) \in (\mathbb{R}^{d},\mathbb{R}^{d})$ sampling from two empirical distributions $(\pi_0,\pi_1)$, the time-dependent probability density function can be denoted as $\rho_t(Z):\mathbb{R}^{d} \times [0,1] \rightarrow \mathbb{R}^{d}$, which satisfies $\rho_0(X^T) = \rho_1(X^S) + \int_{\rho_1(Z)}^{\rho_0(Z)} \partial \rho_t(Z)$. Continuous normalized flows (CNFs)~\citep{CNF} optimize $g_{v_\theta}(\cdot)$ by solving a flow matching problem:
\begin{equation}
\footnotesize
\begin{aligned}
\operatorname*{arg\,min}_{v_\theta} & \int_0^1\mathbb{E}[||\partial \rho_t(Z)/\partial t-g_{v_\theta}(Z_t,t)||]dt.\\
\end{aligned}
\label{eq:rectified_flow_1}
\end{equation}
In inference, the reverse sampling process can be achieved by solving the ordinary differential equation (ODE) $\frac{d\hat{Z}_t}{dt} = -g^*_{v_\theta}(\hat{Z}_t,t)$ through the numerical integration with an initial condition $\hat{Z}_1\sim \pi_1$ and the optimized meta-encoder $g^*_{v_\theta}(\cdot)$, which ultimately yields the synthesized data $\hat{Z}_0$ that is expected to satisfy $\hat{Z}_0\sim \pi_0$. In this context, the trajectory $\{Z_t\}_t$ remains indeterminate in the absence of additional constraints, a factor that often leads to the collapse of the student during training. Drawing inspiration from diffusion models, which have demonstrated exceptional efficacy in image synthesis, the incorporation of prior forward processes (\textit{i.e.}, fixed probability flow paths) is equally essential and advantageous. Therefore, we utilize a noise schedule-like definition~\citep{dpm_solver,kingma2021variational} to model stochastic interpolants~\citep{stochastic_interpolant} ($\alpha_t$, $\sigma_t\in \mathbb{R}^{+}$ are differentiable functions of $t$):
\begin{equation}
\fontsize{8pt}{11pt}\selectfont
\begin{aligned}
Z_t = \alpha_t X^S + \sigma_t X^T,\ s.t.\ \lim_{t\rightarrow 0}\alpha_t = 0, \lim_{t\rightarrow 0}\sigma_t = 1, \lim_{t\rightarrow 1}\sigma_t = 0.\\
\end{aligned}
\label{eq:rectified_flow_2}
\end{equation}
Note that withdrawing constraint $\lim_{t\rightarrow 1}\alpha_t = 1$ is used to ensure that the VE ODE can be unified within our definition. This conversion is convenient since it only needs to modify the initial condition $Z_1$ from $X^S$ to $\alpha_1 X^S$.
\vspace{-3pt}

\paragraph{Noise Schedules.} Different noise schedules affect the effectiveness of knowledge transfer. In this work, we consider three well-known noise schedules, namely VP ODE~\citep{sde}, VE ODE~\citep{sde}, and Rectified flow~\citep{iclr22_rect}, to analyze which noise schedule is most beneficial for knowledge transfer. Among them, VP ODE and VE ODE are derived from VP SDE and VE SDE, respectively~\citep{ddpm_begin,ncsn,sde}. Review these three noise schedules, which can be defined as follows:
\begin{itemize}[topsep=-5pt]
\setlength{\itemsep}{0pt}
\setlength{\parsep}{0pt}
\setlength{\parskip}{0pt}
\item[] \textbf{VP ODE:} $\alpha_t = \textrm{exp}(-\frac{1}{4}a(1-t)^2-\frac{1}{2}b(1-t))$; $\sigma_t = \sqrt{1-\alpha_t^2}$,\quad\textit{s.t. $\mathit{a=19.9}$, $\mathit{b=0.1}$.}
\item[] \textbf{VE ODE:} $\alpha_t =  a(\frac{b}{a})^{t}$; $\sigma_t = 1$,\quad\textit{s.t. $\mathit{a=0.02}$, $\mathit{b=100}$.}
\item[] \textbf{Rectified flow:} $\alpha_t = t$; $\sigma_t = 1-t$.
\end{itemize}

\section{Methodology}

In this section, we first present FM-KT, a novel method for multi-step sampling designed for precise knowledge transfer. Subsequently, a theoretical analysis of the reliability and effectiveness of FM-KT is given. Finally, we further introduce two variants: a lightweight offline knowledge distillation method FM-KT$^\Theta$ and an online knowledge distillation method OFM-KT.

\subsection{Serial Training Paradigm}

The significant challenge in implementing CNFs is the risk of ``cheating''. This risk can hinder the distilled student from acquiring meaningful representations. Our empirical observation indicates that if we directly introduce Rectified Flow, it only achieves an accuracy of 3.42\% with the WRN-40-2-WRN-16-2 pair on CIFAR-100. Furthermore, from a theoretical perspective, the ELBO of CNFs can be written as $\log p_{v_\theta}(Z_0) \geq  -\mathbb{E}_{q(Z_{1/N:1}|Z_0)} [\log\frac{q(Z_{1}|Z_0)}{p_{v_\theta}(Z_1)p_{v_\theta}(Z_0|Z_{1/N})} +\sum_{i=1}^N\log \frac{q(Z_{(i-1)/N}|Z_{i/N},Z_0)}{p_{v_\theta}(Z_{(i-1)/N}|Z_{i/N})}]$. The condition $Z_{i/N}$ of $p_{v_\theta}(\cdot|Z_{i/N})$ in this equation can be obtained by stochastic interpolants, which necessarily incorporate the target information $X^T$ to be learnt, ultimately causing the student to fall into trivial solutions when training converges.

\begin{figure}[t]
\centering
\includegraphics[width=0.95\linewidth,trim={0cm 0.2cm 0cm 0cm},clip]{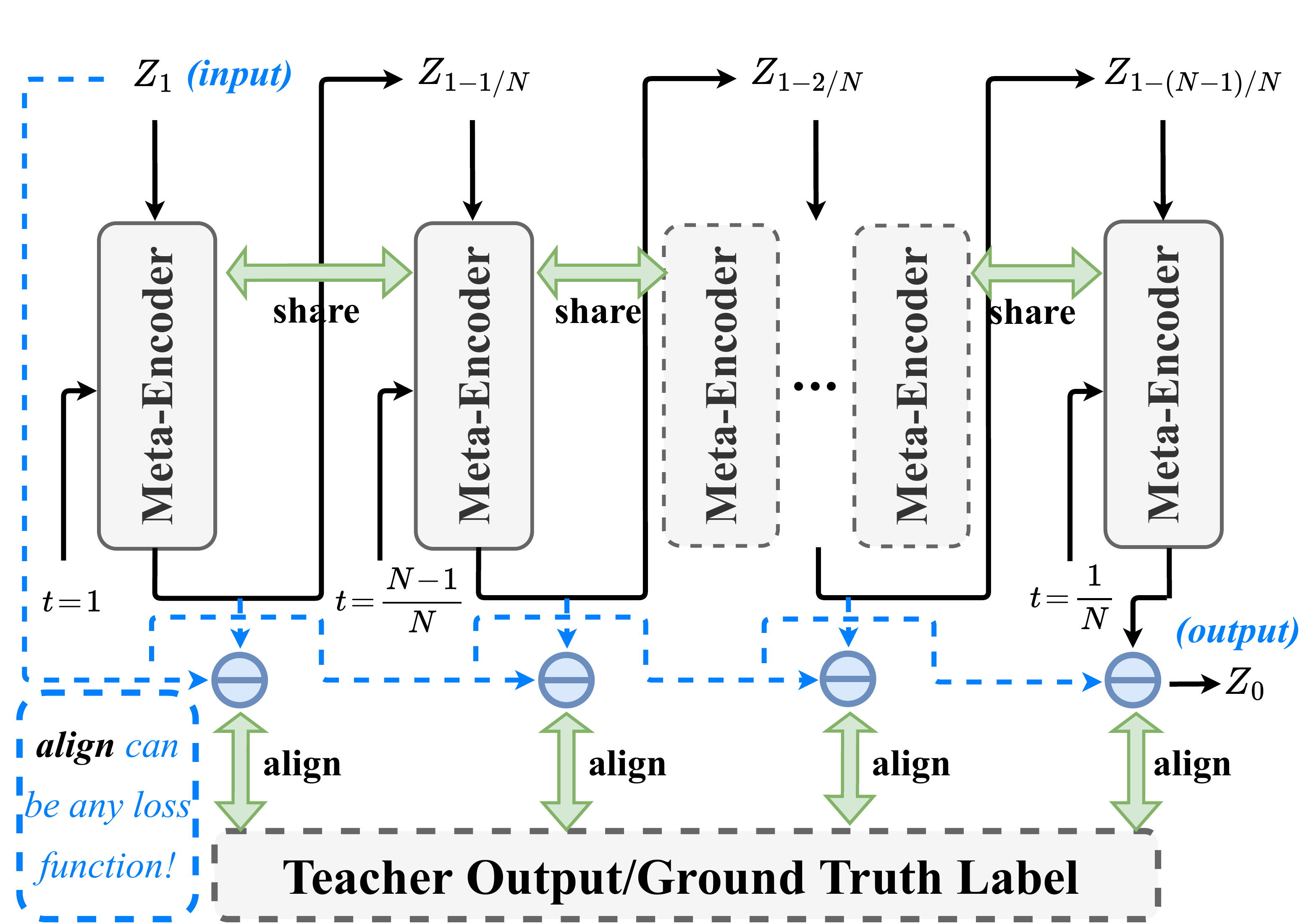}
\caption{The overall structure of FM-KT.}
\label{fig:overall_of_fm_kd}
\vspace{-12pt}
\end{figure}

This requires us to modify the training paradigm of the original CNFs according to the properties of the knowledge transfer scenario. As illustrated in Fig.~\ref{fig:overall_of_fm_kd}, we propose a serial training paradigm within FM-KT to address ``cheating'', which can be denoted as (see Appendix~\ref{apd:unified} for derivation)\begin{equation}
\vspace{-2pt}
\fontsize{8pt}{11pt}\selectfont
\begin{aligned}
 &\mathcal{L}_\textrm{FM-KT} \!=\! \mathbb{E} [\frac{1}{N}\sum_{i=0}^{N-1} \!L(\mathcal{T}((\nabla_t\alpha_t Z_1 \!\!-\!\! g_{v_\theta}(Z_{1\!-\!i/N},1-i/N))/\!-\!\nabla_t\sigma_t) \\
 & , X^T) + \underbrace{L(\mathcal{T}((\nabla_t\alpha_tZ_1 - g_{v_\theta}(Z_{1-i/N},1-i/N))/-\nabla_t\sigma_t), Y)}_{\textrm{match the ground truth label (optional)}}],\\
& \textrm{where}\quad Z_{1-i/N} = Z_{1-(i-1)/N} \\
&\quad\quad - g_{v_\theta}(Z_{1-(i-1)/N},1-(i-1)/N)/N,\quad s.t.\quad i\geq1.
\end{aligned}
\label{eq:fm_kd_1}
\end{equation}
The expectation in Eq.~\ref{eq:fm_kd_1} is taken with respect to $(X^S,X^T,Y)$, and $L(\cdot,\cdot)$ and $Y$ are \rr{the} metric-based distillation method (\textit{i.e.} the loss function) and the ground truth label, respectively. The initial state of the sampling is $Z_1=\alpha_1X^S$. We define $N$ and $K$ as the number of sampling steps during training and inference, respectively. In our work, different values of $K$ are implemented using skip-step sampling of DDIM~\citep{ddim}. \rr{The pseudo code of FM-KT} can be found in Appendix~\ref{apd:pseudocode}. It is guaranteed that $N$ does not exceed 8 (default $N$ as 8) in our experiments thereby avoiding a significant increase in computational cost. It is important to clarify that not only does the training of FM-KT need to be performed by serial, but the inference also relies on multi-step sampling. Intuitively, FM-KT is an interesting ``time-for-accuracy'' algorithm, which in some ways makes a trade-off between time cost and student performance even in inference.
\begin{figure*}[!t]
\begin{center}
\includegraphics[width=1.0\linewidth,trim={0cm 0.5cm 0cm 0cm},clip]{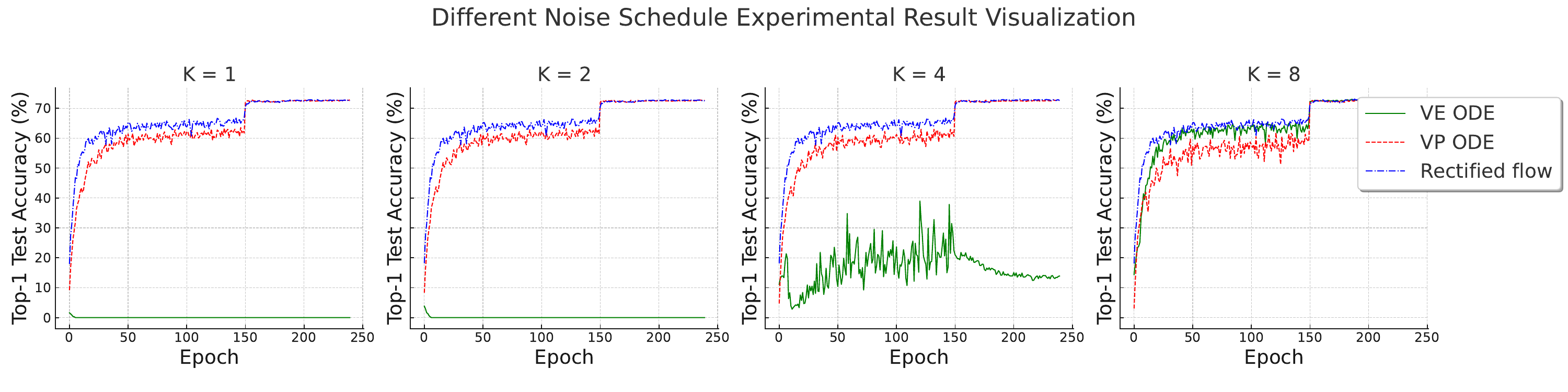}
\vspace{-20pt}
\end{center}
   \caption{Trajectories of Top-1 test accuracy with WRN-40-2-WRN-16-2 pair on CIFAR-100 for various noise schedules: VP ODE, VE ODE, and Rectified flow. Please refer to Appendix~\ref{apd:unified} for more details.}
\label{fig:noise_schedule_main_paper}
\vspace{-10pt}
\end{figure*}

We prove that minimizing $\mathcal{L}_\textrm{FM-KT}$ is closely equivalence to the minimization of the upper bound of the negative log-likelihood of $X^T$ as shown by the following Theorem~\ref{the:training_paradiam}. This theorem leads to efficient training and theoretically guarantees the rationality and practicality of FM-KT.
\begin{theorem}
\label{the:training_paradiam}
(Proof in Appendix~\ref{apd:training_paradiam}) Optimizing $\mathcal{L}_\textrm{FM-KT}$ not only avoids ``cheating'' by accessing $X^T$ during training, but also establishes an equivalence to the upper bound of the negative log-likelihood of $X^T$.
\end{theorem}
\vspace{-5pt}

\subsection{Choice of Noise Schedule} 
Here, we examine two non-straight noise schedules, namely VP ODE and VE ODE, as well as one straight noise schedule referred to as Rectified flow to acquire the most desirable noise schedule in our study. Thus, we empirically conduct experiments with the WRN-40-2-WRN-16-2 pair on CIFAR-100 and present outcomes in Fig.~\ref{fig:noise_schedule_main_paper}. We discover that Rectified flow yields the best stability and effectiveness since the invariance of $\nabla_t \alpha_t$ and $\nabla_t \sigma_t$ at each time point ensures stability during training. Therefore, unless otherwise specified, Rectified flow is applied as the noise schedule by default in all our experiments.

\vspace{-5pt}
\subsection{Serve to Feature-/Logit-based Distillation}
\begin{figure}[t]
\centering
\includegraphics[width=0.9\linewidth,trim={0cm 0.2cm 0cm 0cm},clip]{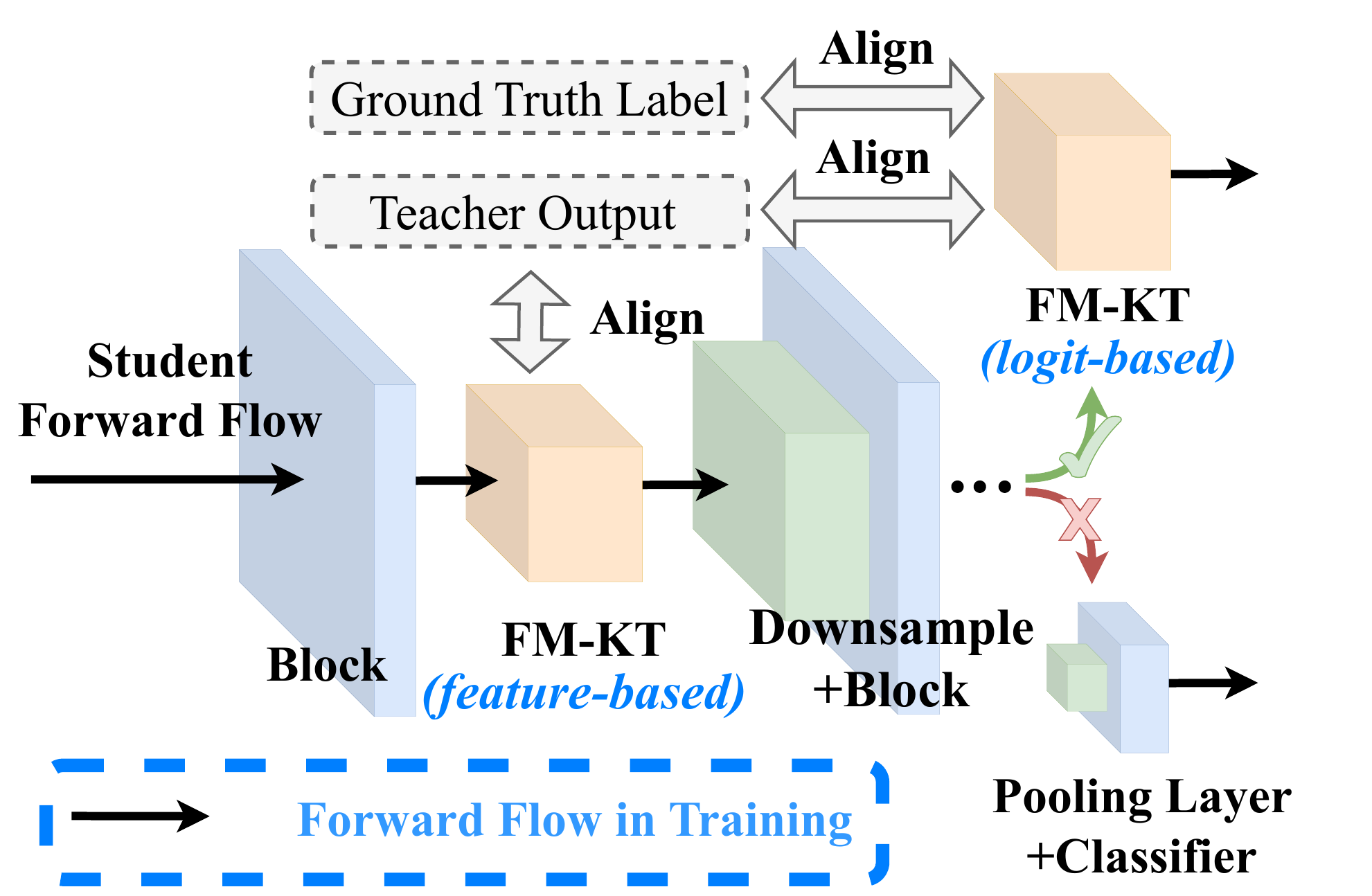}
\caption{An example of FM-KT usage.}
\label{fig:example_fm_kd}
\vspace{-8pt}
\end{figure}
\rr{By simplistically integrating FM-KT into the standard distillation framework, it can serve to the majority of feature-/logit-based distillation algorithms. This introduction is straightforward; it involves replacing the loss function in FM-KT with suitable metric-based distillation approaches.} Practically, FM-KT is strategically placed between different layers of the student to accomplish knowledge transfer. We give \rr{an} example in Fig.~\ref{fig:example_fm_kd}. For feature-based distillation, FM-KT is inserted between the intermediate layers of the student, typically before the downsampling layer. This insertion does not alter the rest of the student architecture. Furthermore, for logit-based distillation, FM-KT replaces the original pooling layer, linear classification layer, or even the penultimate one or two layers (\textit{e.g.} convolution, activation and normalization layers), to achieve logit-level matching. \rr{In our experiments, the unique replacement of the extra penultimate one or two layers is only used for the student on CIFAR-100 is MobileNetV2.} Besides, as shown in Eq.~\ref{eq:fm_kd_1}, FM-KT can optionally add a new loss function by substituting the ground truth label for $X^T$, enabling consistency with the classical logit-based distillation paradigm.

For complex distillation algorithms with learnable encoders, such as MasKD~\citep{Maskd}, we can denote the entire algorithm as a loss function. Hence, it is plausible to replace $L(\cdot,\cdot)$ \rr{in FM-KT} with these algorithms to enable ``serve to feature-/logit-based distillation''. In this study, we focus on simple yet effective metric-based distillation methods, including vanilla KD, DKD, PKD, and DIST. The adaptation of more complex distillation algorithms is earmarked for future work, which will help further ascertain the robust applicability of FM-KT.
\vspace{-5pt}
\subsection{Approximate to Ensemble}
We attribute the ability of FM-KT to efficiently realize knowledge transfer to its multi-step sampling enabled by numerical integration. In the training and inference phases, $N$ ($\leq$ 8) is controlled to be small enough that FM-KT no longer satisfies the form of CNFs. However, in case the noise schedule is set as Rectified flow, Euler's method can be rewritten as averaging multiple time-step outputs, which intuitively approximates an ensemble approach. For completeness, we provide in-depth theoretical support for this through the perspective of error analysis in Proposition~\ref{pps:link_ensemble_with_flow}.
\begin{proposition}
\label{pps:link_ensemble_with_flow}
(Proof in Appendix~\ref{apd:link_ensemble_with_flow}) Assume the noise schedule is set as Rectified flow, FM-KT can be considered a unique implicit ensemble algorithm. The number of outputs used for ensemble is equivalent to the number of samplings.
\end{proposition}
As is well-known, some past methods~\citep{dpm_solver,icml23_consistency} for error analysis in the sampling process of the diffusion model use absolute error bound, which achieves recursion and thus scaling of the \rr{accumulated} error value. We discard the constraint on the absolute value and employ recursion and Taylor expansion in the derivation of Proposition~\ref{pps:link_ensemble_with_flow}. As a result, we obtain the interesting conclusion that the truncation error, which is supposed to be progressively scaled, makes the sampling process of the FM-KT a unique implicit ensemble approach in this proposition.

\subsection{Lightweight FM-KT$^{\Theta}$ without Additional Inference Burden}
The multi-step sampling of FM-KT introduces additional overhead during inference. To facilitate efficient deployment, we propose a streamlined variant FM-KT$^{\Theta}$ for logit-based distillation, which enhances the process by distilling $Z_0$ from FM-KT into the existing classification head (\textit{i.e.} the original student's classification head) $\mathcal{T}_\textrm{vanilla}(\cdot)$, ensuring no extra inference cost. Essentially, this is a concept of progressive distillation, which enhances student performance by effectively reducing the gap between the teacher and the student. During training, we reformulate the loss function to accommodate this integration:
\begin{equation}
\fontsize{8pt}{11pt}\selectfont
\begin{aligned}
 & \mathcal{L}_{\textrm{FM-KT}^{\Theta}} = \\
 & \mathbb{E} [L(\mathcal{T}_\textrm{vanilla}(X^S),\mathcal{T}(Z_0)) + \alpha^{\Theta}L(\mathcal{T}_\textrm{vanilla}(X^S),Y)] + \mathcal{L}_\textrm{FM-KT},\\
\end{aligned}
\label{eq:fm_kd_lightweight}
\end{equation}
where the expectation is taken with respect to $(X^S,X^T,Y)$, and $\alpha^{\Theta}$ refers to the balance weight. In inference, we can directly utilize $\mathcal{T}_\textrm{vanilla}(\cdot)$ to achieve prediction without going through $g_{v_\theta}(\cdot)$ and $\mathcal{T}(\cdot)$ to increase the sampling burden.

\subsection{Translate to Online Knowledge Distillation}
Numerous Online Knowledge Distillation (Online KD) algorithms essentially integrate the outputs of multiple branches, thus avoiding asynchronous updating of gradients and ultimately improving the generalization ability of the student. FM-KT and Online KD have different approaches but equally satisfactory results, which provides the feasibility for FM-KT to be converted to Online KD. \rr{In comparison to Offline Knowledge Distillation (Offline KD), Online KD doesn't use an explicit teacher; instead, the teacher is represented by a weighted average of branches in the student.} Similarly, we can achieve the goal ``translate to Online KD'' by simply replacing $X^T$ in Eq.~\ref{eq:fm_kd_1} with the final result after sampling with Euler's method. In detail, we first obtain the sampling result $Z_0$ by continuously calling Euler's method $Z_{1-i/N}\!=\!Z_{1\!-\!(i\!-\!1)/N}\!-\!g_{v_\theta}(Z_{1\!-\!(i\!-\!1)/N},1\!-\!(i\!-\!1)/N)/N$. Finally, we retain the portion of FM-KT that matches the ground truth label and add the \rr{Online KD} loss to it
\begin{equation}
\vspace{-14pt}
\fontsize{8pt}{11pt}\selectfont
\begin{aligned}
&\mathcal{L}_\textrm{OFM-KT} = \mathbb{E}_{(X^S,Y)} [\frac{1}{N}\sum_{i=0}^{N-1} \\
&\quad\underbrace{L(\mathcal{T}((\nabla_t\alpha_t Z_1 - g_{v_\theta}(Z_{1-i/N},1-i/N))/-\nabla_t\sigma_t), Z_0)}_{\textrm{the Online KD loss}} \\
 &+\underbrace{L(\mathcal{T}((\nabla_t\alpha_t Z_1 - g_{v_\theta}(Z_{1-i/N},1-i/N))/-\nabla_t\sigma_t), Y)}_{\textrm{match the ground truth label}}]. \\
\end{aligned}
\label{eq:fm_kd_2}
\end{equation}

The variant $\mathcal{L}_\textrm{OFM-KT}$ can be empirically understood as a novel Online KD algorithm OFM-KT. Compared with traditional Online KD algorithms including ONE~\citep{ONE}, KDCL~\citep{PCLs} and AHBF-OKD~\citep{AHBF-OKD}, OFM-KT has some unique characteristics, including the meta-encoder shares parameters at different time points, whereas traditional Online KD algorithms do not. Besides, the input of the meta-encoder in OFM-KT is different at different time points, and as $t\rightarrow0$, the input contains more target information. In contrast, the traditional Online KD has the same input for each branch. This means that OFM-KT achieves ensemble through various inputs instead of unshared parameters.

\begin{table*}[!t]
\begin{center}
\resizebox{0.94\textwidth}{!}{%
\begin{tabular}{c|cccccccc}
\hline
\makecell{Teacher\\Student} & \makecell{Meta-\\Encoder} & \makecell{ResNet56\\ResNet20} &\makecell{WRN-40-2\\WRN-16-2}&\makecell{WRN-40-2\\WRN-40-1}&\makecell{ResNet32$\times$4\\ResNet8$\times$4}&\makecell{VGG13\\VGG8}&\makecell{VGG13\\MobileNetV2}&\makecell{WRN-40-2\\ShuffleNetV1}\\
\hline
Teacher & \xmark & 73.24 & 75.61 & 75.61 & 79.42 & 74.64 & 75.61 & 75.61 \\
\baseline{Student}& \baseline{\xmark} & \baseline{69.06}& \baseline{73.26}&\baseline{71.98}&\baseline{72.50}&\baseline{70.36}&\baseline{64.60}&\baseline{70.50}\\
\hline
\noalign{\smallskip}
ATKD&\xmark & 70.55&74.08&72.77&73.44&71.43&59.40&72.73\\ 
SPKD&\xmark &69.67&73.83&72.43&72.94&72.68&66.30&74.52\\
CRD&\xmark &71.16&75.48&74.14&75.51&73.94& 69.73 &76.05\\
vanilla KD&\xmark &70.66&74.92&73.54&73.33&	72.98&67.37&74.83\\
DKD&\xmark &71.97&76.24&74.81&76.32& 74.68 &69.73 &76.70\\
DIST&\xmark &71.26&75.29&74.42&75.79&73.11&68.48&75.23\\
FM-KT$^{\Theta}$&\xmark & 72.20 & 75.98 & 74.99 & 76.52 & 74.82 & 69.90 & 77.19 \\\hline\noalign{\smallskip}
DiffKD&\cmark &71.92& 76.13 &74.09&76.31& - & - & - \\
FM-KT ($K$=1)&\cmark & 74.28&77.14&75.88&76.74& 75.21 &69.68 &76.34\\
FM-KT ($K$=2)&\cmark &74.09&76.58&74.52&74.98& 74.86&69.52&75.55\\
FM-KT ($K$=4)&\cmark & \textbf{75.12}&77.69&\textbf{76.24}&77.49& 75.42& \textbf{69.94}&76.95\\
FM-KT ($K$=8)&\cmark & 74.97&\textbf{77.84}&76.09&\textbf{77.71}&\textbf{75.46}&\textbf{69.94}&\textbf{77.21}\\\hline
\end{tabular}}
\vspace{-10pt}
 \end{center}
\caption{Results of different Offline KD methods on CIFAR-100. Among them, ATKD, SPKD, CRD and DiffKD belong to feature-based distillation, while vanilla KD, DKD and DIST belong to logit-based distillation.}
\label{table:cifar_results}
\end{table*}
\begin{table*}[t!]
\begin{center}
\resizebox{0.999\textwidth}{!}{%
\begin{tabular}{cccc|ccccc|cccccc}
\hline
T-S Pair & Acc. & Tea. & \baseline{Stu.} & \makecell{Vanilla KD} & \makecell{ReviewKD} & \makecell{DKD} & \makecell{DIST} & FM-KT$^{\Theta}$ & \makecell{DiffKD} & \makecell{KDiffusion} & \makecell{FM-KT\\($K$=1)} & \makecell{FM-KT\\($K$=2)} & \makecell{FM-KT\\($K$=4)} & \makecell{FM-KT\\($K$=8)} \\
Meta-encoder & - & - & - & \xmark & \xmark & \xmark & \xmark & \xmark & \cmark & \cmark & \cmark & \cmark & \cmark & \cmark \\
\hline
\multirow{2}{*}{R34-R18} & Top-1 & 73.31 & \baseline{69.75} & 70.66 & 71.61 & 71.70 & 72.07 & 72.14 & 72.49 & 72.04 & 72.49 & 72.86 & 73.08 & \textbf{73.17} \\ 
&Top-5 & 91.42 & \baseline{89.08} & 89.88 & 90.51 & 90.41& 90.42 & 90.44 & 90.71 & 90.53 & 
90.83 & 91.00 & 91.12 & \textbf{91.18} \\ 
\multirow{2}{*}{R50-MBV1} & Top-1 & 76.16 & \baseline{70.13} & 70.68 & 72.56 & 72.05 & 73.24 & 73.29 & 73.78 & 73.62 & 73.61 & 74.01 & 74.20 & \textbf{74.22} \\   
& Top-5 & 92.86 & \baseline{89.49} & 90.30 & 91.00 & 91.05 & 91.12 & 91.15 & 91.48 & 91.82 & 91.36 & 91.71 & \textbf{91.84} & 91.81 \\\hline
\end{tabular}}
\vspace{-10pt}
 \end{center}
\caption{Results of different Offline KD methods on ImageNet-1k. ``R34-R18'' and ``R50-MBV1'' refer to ``ResNet34-ResNet18 pair'' and ``ResNet50-MobileNetV1 pair'', respectively.}
\label{table:imagenet_results}
\vspace{-5pt}
\end{table*}
\section{Experiments}
We perform comparison and ablation experiments on CIFAR-100, ImageNet-1k and MS-COCO. The implementation details of FM-KT, FM-KT$^{\Theta}$, and OFM-KT can be found in Appendix~\ref{apd:implementation_detail}, and the experimental result of MS-COCO can be found in Appendix~\ref{apd:obj}. Note that all normalization layers in the meta-encoder are not BatchNorm, because their inputs are various at different time points, so the statistics of the mean and variance will encounter difficulties, thereby causing training collapse. Moreover, we introduce a strategy named pair decoupling (PD), which is controlled by the hyperparameter dirac ratio $\beta_d$, applied to shuffle part of the sample pairs in a batch. This approach is particularly effective for feature-based distillation in image classification tasks, and its detailed description and specific implementation can be found in Appendix~\ref{apd:pair_decoupling} and~\ref{apd:pseudocode}, respectively.\begin{table*}[!t]
\begin{center}
\resizebox{0.8\textwidth}{!}{%
\begin{tabular}{c|cccccc}
\hline
\makecell{Architecture} & Meta-encoder & {ResNet32} & {ResNet110}&{VGG16}& {DenseNet40-2}& {MobileNetV2}\\
\hline
\baseline{Student}&\baseline{\xmark} &\baseline{71.28}& \baseline{76.21}&\baseline{74.32}&\baseline{71.03}&\baseline{59.79}\\\hline
CL &\baseline{\cmark} & 72.33 & 78.83 & 74.33 & 71.45 & 60.63 \\
ONE &\baseline{\cmark} & 72.45 & 78.44 & 74.38 & 71.39 & 60.84 \\
FFSD-C &\baseline{\cmark} & 74.50 & 78.83 & 74.89 & 71.74 & 61.88 \\
ABHF-OKD &\baseline{\cmark} & \textbf{74.81} & 79.04 & 75.08 & 72.12 & 62.23 \\
OFM-KT ($K$=1)&\baseline{\cmark} & 72.86 & 79.49 & 75.07 & 73.12 & 63.62 \\
OFM-KT ($K$=2)&\baseline{\cmark} & 73.02 & \textbf{79.50} &\textbf{ 75.10} & 73.34 & \textbf{63.67} \\
OFM-KT ($K$=4)&\baseline{\cmark} & 73.10 & 79.45 & 75.09 & \textbf{73.40} & 63.63 \\
OFM-KT ($K$=8)&\baseline{\cmark} & 73.07 & 79.47 & 75.06 & 73.39 & 63.61 \\\hline
\end{tabular}}
\vspace{-10pt}
 \end{center}
\caption{Results of different Online KD methods on CIFAR-100. The metric is the Top-1 accuracy.}
\label{table:cifar_results_online}
\vspace{-5pt}
\end{table*}\begin{table*}[!t]
\begin{center}
\resizebox{1.\textwidth}{!}{%
\begin{tabular}{cc|cccccccc}
\hline
\vspace{1pt}
Architecture & \baseline{Student} & ONE & OKDDip & FFSD-C & ABHF-OKD & OFM-KT ($K$=1) & OFM-KT ($K$=2) & OFM-KT ($K$=4) & OFM-KT ($K$=8) \\
Meta-encoder & \xmark & \cmark & \cmark & \cmark & \cmark & \cmark & \cmark & \cmark & \cmark \\
\hline
\multirow{1}{*}{ResNet18} & \baseline{69.75} & 70.55& 70.63 & 70.15 & 70.72 & 71.38 & 71.52 & \textbf{71.56} & \textbf{71.56} \\
\multirow{1}{*}{ResNet34} & \baseline{73.24} & 74.10 & 74.40 & 74.20 & \textbf{74.53} & 74.16 & 74.20 & 74.20 & 74.20\\\hline
\end{tabular}}
\vspace{-10pt}
 \end{center}
\caption{Results of different Online KD methods on ImageNet-1k. The metric is the Top-1 accuracy.}
\label{tab:imagenet_results_online}
\vspace{-5pt}
\end{table*} \rr{The impact of the normalization layer selection in the meta-encoder, where stages used for distillation in the feature-based scenario, and ideal configuration of dirac ratio $\beta_d$ can be found in the additional ablation experiments in Appendix~\ref{apd:add_ablation}.} By default, we set $\beta_d$ as 0.25 and use the 1st and 2nd last stages for feature-based distillation in image classification tasks.

Crucially, detailed results about \textbf{the stronger teacher comparison, vision transformer comparison, and visualization of sampling trajectory}, can be found in Appendix~\ref{sec:stronger},~\ref{sec:deit} and~\ref{apd:visualization}, respectively.
\begin{figure*}[!t]
\begin{center}
\includegraphics[width=0.9\linewidth]{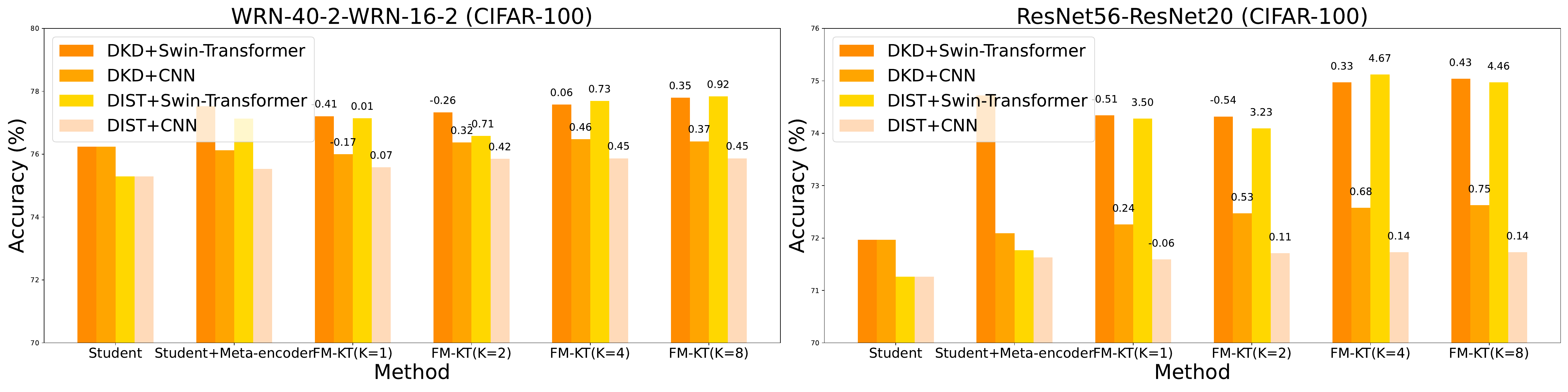}
\vspace{-10pt}
\end{center}
   \caption{Results of experiments on the ensemble capabilities of FM-KT on CIFAR-100. The numbers on the bars represent their performance gains compared to Student+Meta-encoder.}
\label{fig:effective}
\vspace{-10pt}
\end{figure*}
\begin{figure*}[!t]
\begin{center}
\includegraphics[width=0.95\linewidth]{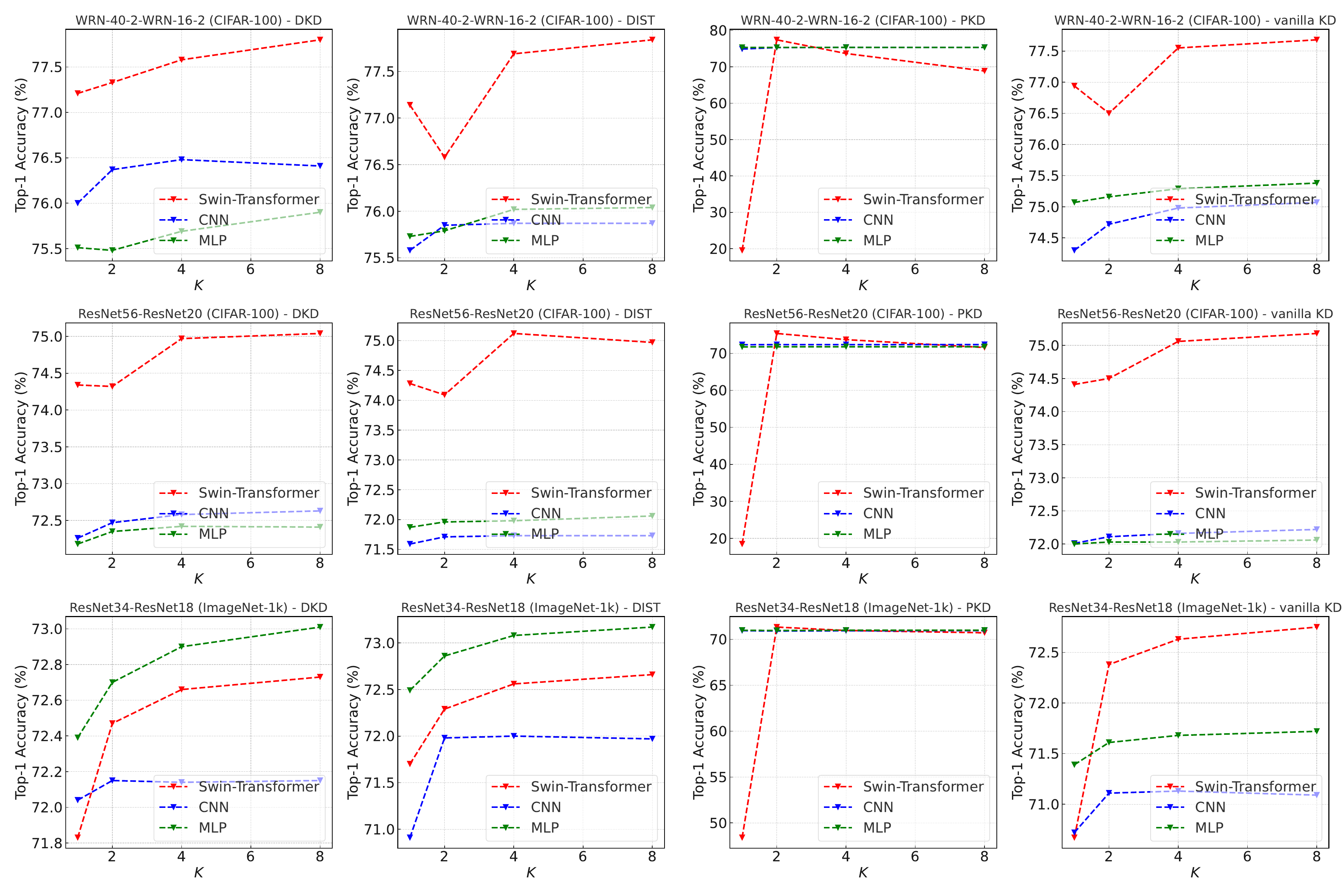}
\vspace{-10pt}
\end{center}
   \caption{Ablation results about the loss function and the meta-encoder on CIFAR-100 and ImageNet-1k.}
\label{table:ablation_loss_encoder}
\vspace{-5pt}
\end{figure*}
\vspace{-5pt}
\subsection{Image Classification Comparison}
\paragraph{Offline Knowledge Distillation.} On CIFAR-100, we conduct experiments on teacher-student pairs including ResNet56-ResNet20, WRN-40-2-WRN-16-2~\citep{Zagoruyko2016WRN}, WRN-40-2-WRN-40-1, ResNet32$\times$4-ResNet8$\times$4, VGG13-VGG8~\citep{Inceptionv1}, VGG13-MobileNetV2~\citep{mobilenetv2} and WRN-40-2-ShuffleNetV1~\citep{shufflenet} pairs. We compare FM-KT with state-of-the-art methods including ATKD~\citep{ATKD}, SPKD, CRD, DiffKD, KDiffusion, vanilla KD, DKD and DIST, and present the results in Table~\ref{table:cifar_results}. As shown in Table~\ref{table:cifar_results}, FM-KT significantly outperforms prior KD methods with all pairs. Note that FM-KT improves the student performance on ResNet56-ResNet20, WRN-40-2-WRN-16-2, WRN-40-2-WRN-40-1 and VGG13-VGG8 pairs by 3.15\%, 1.60\%, 1.43\% and 0.64\%, respectively, compared with the best prior methods. Moreover, our lightweight variant FM-KT$^\Theta$, which without additional computational cost in inference, achieves state-of-the-art performance across a wide range of teacher-student pairs. On ImageNet-1k, FM-KT treats DIST as its $L(\cdot,\cdot)$ (\textit{w.r.t.} baseline). Compared with DIST, FM-KT exceeds DIST on ResNet34-ResNet18 and ResNet50-MobileNetV1 by 1.10\% and 0.98\%, respectively. Specifically, FM-KT demonstrates superior performance over similar algorithms, DiffKD and KDiffusion, showing a substantial margin of improvement on both ResNet34-ResNet18 and ResNet50-MobileNetV1 pairs. However, it should be noted that DiffKD introduces 11 additional convolutional layers in its encoder (considering its mentioned Diffusion Model and Noise Adapter), while in contrast, FM-KT employs 2-layer MLP with only 4 linear layers as its meta-encoder. Furthermore, the lightweight FM-KT$^\Theta$ outperforms all algorithms with no additional computational overhead in inference, validating its effectiveness and applicability.

\vspace{-8pt}
\paragraph{Online Knowledge Distillation.} We present the results of the comparison between OFM-KT and prior state-of-the-art approaches CL~\citep{CL}, ONE~\citep{ONE}, OKDDip~\citep{OKDDIP}, FFSD-C~\citep{FFSD} and ABHF-OKD~\citep{AHBF-OKD} in Table~\ref{table:cifar_results_online} and~\ref{tab:imagenet_results_online}. Among them, Table~\ref{table:cifar_results_online} illustrates the experimental results on CIFAR-100. We can observe OFM-KT beats all comparison methods on \rr{ResNet110, VGG16, DenseNet40-2, and MobileNetV2}. For the results on ImageNet-1k on Table~\ref{tab:imagenet_results_online}, OFM-KT outperforms other methods on ResNet18, albeit lagging behind the optimal ABHF-OKD by a marginal 0.33\% on ResNet34. Importantly, regarding both ResNet18 and ResNet34, OFM-KT necessitates merely two Number of Function Evaluations (NFEs) to attain the best results. This \rr{indicates} that OFM-KT corresponds to Online KD, which is the aggregated outcome of two branches sharing parameters. Hence, this compellingly substantiates that OFM-KT is a potent Online KD algorithm.

\subsection{Ablation Studies}
\label{sec:ablation_studies}
\noindent{\bf The Number of Sampling Steps $K$ in Inference.}
\begin{figure}[t]
\centering
\includegraphics[width=0.95\linewidth,trim={0cm 0.2cm 0cm 0cm},clip]{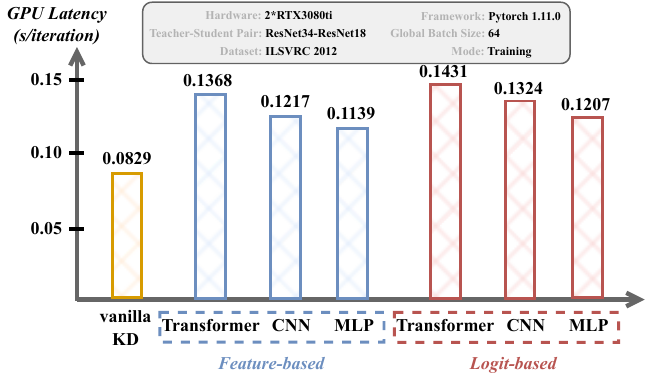}
\vspace{-5pt}
\caption{GPU latency comparison between FM-KT and vanilla KD during training. The computational cost comparison and analysis in inference can be found in Appendix~\ref{apd:add_training_and_inference_cc_discussion}.}
\label{fig:gpu_latency_of_fm_kd}
\vspace{-12pt}
\end{figure}
We can also call $K$ \rr{as} NFEs, an important metric affecting the GPU latency during inference. Both FM-KT and OFM-KT have a similarity form to the diffusion models family (\textit{e.g.} VE-SDE, VP-SDE, EDM~\citep{nips22_design}) and INN~\citep{INN}, in that after obtaining the training weights, the NFEs can be modified at the time of inference to trade-off effective and efficiency. We can see from Table~\ref{table:cifar_results},~\ref{table:cifar_results_online},~\ref{table:imagenet_results},~\ref{tab:imagenet_results_online} and Fig.~\ref{table:ablation_loss_encoder} that increasing $K$ will improve the student performance, but in general $K\!=\!2$ will achieve quite satisfactory results. Note that the combination of PKD and Swin-Transformer on ImageNet-1k in Fig.~\ref{table:ablation_loss_encoder} has a large difference in the results achieved by the different NFEs, but the best results are superior to the combination of PKD and MLP/CNN. This might be because Swin-Transformer does not have inductive bias~\citep{howvitwork} and the specificity that PKD is a feature-based distillation method.

\noindent{\bf The Computational Cost Analysis.} For computational burden during training, as illustrated in Fig.~\ref{fig:gpu_latency_of_fm_kd}, applying a serial loss calculation $\mathcal{L}_\textrm{FM-KT}$ does not introduce excessive GPU latency during training. Even the most computationally demanding meta-encoder Transformer has less than double the GPU latency compared to vanilla KD. Additionally, the computational overhead comparison between FM-KT and DiffKD during inference can be found in Appendix~\ref{apd:add_training_and_inference_cc_discussion}.

\noindent{\bf The Effectiveness of Optimization Objective.} We present the validity of the FM-KT optimization objective in Fig.~\ref{fig:effective} to prevent misinterpretations due to the properties of the meta-encoder and the loss function themselves. As $K$ rises, we can observe that the performance improvement becomes increasingly clear. Note that 4.67\% enhancement produced by the DIST+Swin-Transformer on the ResNet56-ResNet20 pair demonstrates that the characteristic of FM-KT \mbox{--} implicit ensemble \mbox{--} does result in performance gains.

\noindent{\bf The Ablation about the Loss Function and Meta-Encoder.} The outcomes of this ablation study are summarized in Fig.~\ref{table:ablation_loss_encoder}. For the meta-encoder, it is demonstrated that the Swin-Transformer yields the most favorable results when combined with any loss function on CIFAR-100. Conversely, on ImageNet-1k, the amalgamation of MLP with DKD, DIST and PKD demonstrates superior performance. Moreover, for the loss function, DIST and DKD exhibit comparable and enhanced performance relative to PKD and vanilla KD across all student-teacher \rr{pairs}.

\section{Related Work}
\paragraph{Knowledge distillation.} The main strategies of knowledge distillation are categorized into three: feature-based~\citep{CRD,FFSD,FKD}, logit-based~\citep{DIST,DKD,FastKD}, and data-based distillations~\citep{nips_kd_aug_statist,TST}. Regardless of the approach, the knowledge transfer framework plays an important role in it. Thus, this paper aims to design a more desirable knowledge transfer framework that can serve both feature-based distillation as well as logit-based distillation.
\vspace{-8pt}
\paragraph{Continuous Network Representation.} There are a number of architectures belonging to continuous network representation, such as RNN~\citep{RNN}, LSTM~\citep{LSTM}, Neural ODE~\citep{NODE}, GflowNet family~\citep{gflownet1,gflownet2}, diffusion model family~\citep{nips22_design,ddpm_begin,ddim}, INN~\citep{INN}, DiffKD~\citep{DiffKD} and KDiffusion~\citep{kdiffusion}. More discussion can be found in Appendix~\ref{apd:cnr}.
\section{Conclusion}
We have proposed a highly scalable framework FM-KT, its lightweight variant FM-KT$^{\Theta}$, and its online knowledge distillation variant OFM-KT for knowledge transfer. The design flexibility of FM-KT, FM-KT$^{\Theta}$ and OFM-KT allows them to be formulated utilizing a loss function of any form and a meta-encoder with any available architecture, making them adaptable for progressive transformation focused both on features and logits via arbitrary noise schedules. Theoretically, we have proven that the optimization objective of FM-KT is equivalent to minimizing the upper bound of the negative log-likelihood of the target (\textit{e.g.} the teacher's output). In future work, we aim to further explore the design space of FM-KT and extend its application to a broader scope of downstream tasks.

\clearpage

\paragraph{Impact Statement.} 
Our proposed FM-KT enhances computational efficiency in lightweight models on edge devices, enabling entities with limited computational resources, such as certain companies and laboratories, to effectively leverage knowledge from large-scale models. However, FM-KT also entails modifications to the student model's architecture, posing potential challenges under certain deployment circumstances that are constrained by hardware specificity.

\nocite{langley00}

\bibliography{ref}

\begin{thebibliography}{62}
\providecommand{\natexlab}[1]{#1}
\providecommand{\url}[1]{\texttt{#1}}
\expandafter\ifx\csname urlstyle\endcsname\relax
  \providecommand{\doi}[1]{doi: #1}\else
  \providecommand{\doi}{doi: \begingroup \urlstyle{rm}\Url}\fi

\bibitem[Achiam et~al.(2023)Achiam, Adler, Agarwal, Ahmad, Akkaya, Aleman, Almeida, Altenschmidt, Altman, Anadkat, et~al.]{GPT4}
Achiam, J., Adler, S., Agarwal, S., Ahmad, L., Akkaya, I., Aleman, F.~L., Almeida, D., Altenschmidt, J., Altman, S., Anadkat, S., et~al.
\newblock Gpt-4 technical report.
\newblock \emph{arXiv preprint arXiv:2303.08774}, 2023.

\bibitem[Ahn et~al.(2019)Ahn, Hu, Damianou, Lawrence, and Dai]{VID}
Ahn, S., Hu, S.~X., Damianou, A.~C., Lawrence, N.~D., and Dai, Z.
\newblock Variational information distillation for knowledge transfer.
\newblock In \emph{Computer Vision and Pattern Recognition}, pp.\  9163--9171, Long Beach, CA, USA, Jun. 2019. Computer Vision Foundation / {IEEE}.

\bibitem[Albergo \& Vanden-Eijnden(2022)Albergo and Vanden-Eijnden]{stochastic_interpolant}
Albergo, M.~S. and Vanden-Eijnden, E.
\newblock Building normalizing flows with stochastic interpolants.
\newblock In \emph{International Conference on Learning Representations}, 2022.

\bibitem[Bengio et~al.(2021)Bengio, Jain, Korablyov, Precup, and Bengio]{gflownet1}
Bengio, E., Jain, M., Korablyov, M., Precup, D., and Bengio, Y.
\newblock Flow network based generative models for non-iterative diverse candidate generation.
\newblock In \emph{Neural Information Processing Systems}, pp.\  27381--27394, Virtual Event, Dec. 2021.

\bibitem[Brown et~al.(2020)Brown, Mann, Ryder, Subbiah, Kaplan, Dhariwal, Neelakantan, Shyam, Sastry, Askell, et~al.]{GPT3}
Brown, T., Mann, B., Ryder, N., Subbiah, M., Kaplan, J.~D., Dhariwal, P., Neelakantan, A., Shyam, P., Sastry, G., Askell, A., et~al.
\newblock Language models are few-shot learners.
\newblock \emph{Advances in neural information processing systems}, 33:\penalty0 1877--1901, 2020.

\bibitem[Cao et~al.(2022)Cao, Zhang, Gao, Cheng, Cheng, and Cheng]{PKD}
Cao, W., Zhang, Y., Gao, J., Cheng, A., Cheng, K., and Cheng, J.
\newblock {PKD:} general distillation framework for object detectors via pearson correlation coefficient.
\newblock In \emph{Neural Information Processing Systems}, New Orleans, LA, USA, Dec. 2022.

\bibitem[Chen et~al.(2020)Chen, Mei, Wang, Feng, and Chen]{OKDDIP}
Chen, D., Mei, J.-P., Wang, C., Feng, Y., and Chen, C.
\newblock Online knowledge distillation with diverse peers.
\newblock \emph{Association for the Advance of Artificial Intelligence}, 34\penalty0 (04):\penalty0 3430--3437, 2020.

\bibitem[Chen et~al.(2018)Chen, Rubanova, Bettencourt, and Duvenaud]{NODE}
Chen, R.~T., Rubanova, Y., Bettencourt, J., and Duvenaud, D.~K.
\newblock Neural ordinary differential equations.
\newblock \emph{Neural Information Processing Systems}, 31, 2018.

\bibitem[Chuanguang et~al.(2021)Chuanguang, Zhulin, Linhang, and Yongjun]{HSAKD}
Chuanguang, Y., Zhulin, A., Linhang, C., and Yongjun, X.
\newblock Hierarchical self-supervised augmented knowledge distillation.
\newblock In \emph{International Joint Conference on Artificial Intelligence}, pp.\  1217--1223, Virtual Event, Aug. 2021. IJCAI.

\bibitem[Contributors(2020)]{mmcls}
Contributors, M.
\newblock Openmmlab's image classification toolbox and benchmark.
\newblock \url{https://github.com/open-mmlab/mmclassification}, 2020.

\bibitem[Cubuk et~al.(2020)Cubuk, Zoph, Shlens, and Le]{RandAugment}
Cubuk, E.~D., Zoph, B., Shlens, J., and Le, Q.~V.
\newblock Randaugment: Practical automated data augmentation with a reduced search space.
\newblock In \emph{Proceedings of the IEEE/CVF conference on computer vision and pattern recognition workshops}, pp.\  702--703, 2020.

\bibitem[Gong et~al.(2023)Gong, Lin, Zhang, Shen, Li, Qiao, Ren, Li, Yu, and Ma]{AHBF-OKD}
Gong, L., Lin, S., Zhang, B., Shen, Y., Li, K., Qiao, R., Ren, B., Li, M., Yu, Z., and Ma, L.
\newblock Adaptive hierarchy-branch fusion for online knowledge distillation.
\newblock In \emph{Association for the Advancement of Artificial Intelligence}, volume~37, pp.\  7731--7739, Washington, DC, USA, Jun. 2023. AAAI.

\bibitem[Gou et~al.(2021)Gou, Yu, Maybank, and Tao]{kdsurvey}
Gou, J., Yu, B., Maybank, S.~J., and Tao, D.
\newblock Knowledge distillation: A survey.
\newblock \emph{International Journal of Computer Vision}, 129\penalty0 (6):\penalty0 1789--1819, 2021.

\bibitem[Guo et~al.(2020)Guo, Wang, Wu, Yu, Liang, Hu, and Luo]{PCLs}
Guo, Q., Wang, X., Wu, Y., Yu, Z., Liang, D., Hu, X., and Luo, P.
\newblock Online knowledge distillation via collaborative learning.
\newblock In \emph{Computer Vision and Pattern Recognition}, pp.\  11020--11029, 2020.

\bibitem[Hinton et~al.(2015)Hinton, Vinyals, and Dean]{vanillakd}
Hinton, G., Vinyals, O., and Dean, J.
\newblock Distilling the knowledge in a neural network, 2015.
\newblock URL \url{https://arxiv.org/abs/1503.02531}.

\bibitem[Ho et~al.(2020)Ho, Jain, and Abbeel]{ddpm_begin}
Ho, J., Jain, A., and Abbeel, P.
\newblock Denoising diffusion probabilistic models.
\newblock In \emph{Neural Information Processing Systems}, pp.\  6840--6851, Virtual Event, Dec. 2020. NIPS.

\bibitem[Huang et~al.(2023)Huang, Zhang, Zheng, You, Wang, Qian, and Xu]{DiffKD}
Huang, T., Zhang, Y., Zheng, M., You, S., Wang, F., Qian, C., and Xu, C.
\newblock Knowledge diffusion for distillation.
\newblock \emph{arXiv preprint arXiv:2305.15712}, 2023.

\bibitem[Huang et~al.(2022)Huang, Peng, Dong, Wei, Jiao, and Ye]{GSD}
Huang, W., Peng, Z., Dong, L., Wei, F., Jiao, J., and Ye, Q.
\newblock Generic-to-specific distillation of masked autoencoders.
\newblock In \emph{Computer Vision and Pattern Recognition}, New Orleans, LA, USA, Jun. 2022.

\bibitem[Karras et~al.(2022)Karras, Aittala, Aila, and Laine]{nips22_design}
Karras, T., Aittala, M., Aila, T., and Laine, S.
\newblock Elucidating the design space of diffusion-based generative models.
\newblock \emph{arXiv preprint arXiv:2206.00364}, 2022.

\bibitem[Kingma et~al.(2021)Kingma, Salimans, Poole, and Ho]{kingma2021variational}
Kingma, D., Salimans, T., Poole, B., and Ho, J.
\newblock Variational diffusion models.
\newblock \emph{Neural Information Processing Systems}, 34:\penalty0 21696--21707, 2021.

\bibitem[Krizhevsky et~al.(2009)Krizhevsky, Hinton, et~al.]{CIFAR}
Krizhevsky, A., Hinton, G., et~al.
\newblock Learning multiple layers of features from tiny images.
\newblock 2009.

\bibitem[Lan et~al.(2018)Lan, Zhu, and Gong]{ONE}
Lan, X., Zhu, X., and Gong, S.
\newblock Knowledge distillation by on-the-fly native ensemble.
\newblock In \emph{Neural Information Processing Systems}, pp.\  7528--7538, Montréal Canada, Dec. 2018. MIT Press.

\bibitem[Lee et~al.(2023)Lee, Kim, and Ye]{icml23_curvature}
Lee, S., Kim, B., and Ye, J.~C.
\newblock Minimizing trajectory curvature of ode-based generative models.
\newblock \emph{arXiv preprint arXiv:2301.12003}, 2023.

\bibitem[Li et~al.(2022)Li, Lin, Wang, Wu, Tian, Shao, and Ji]{FFSD}
Li, S., Lin, M., Wang, Y., Wu, Y., Tian, Y., Shao, L., and Ji, R.
\newblock Distilling a powerful student model via online knowledge distillation.
\newblock \emph{IEEE Transactions on Neural Networks and Learning Systems}, pp.\  1--10, 2022.

\bibitem[Lin et~al.(2014)Lin, Maire, Belongie, Hays, Perona, Ramanan, Doll{\'a}r, and Zitnick]{COCO}
Lin, T.-Y., Maire, M., Belongie, S., Hays, J., Perona, P., Ramanan, D., Doll{\'a}r, P., and Zitnick, C.~L.
\newblock Microsoft coco: Common objects in context.
\newblock In \emph{European Conference on Computer Vision}, pp.\  740--755. Springer, 2014.

\bibitem[Lipman et~al.(2022)Lipman, Chen, Ben-Hamu, Nickel, and Le]{CNF}
Lipman, Y., Chen, R.~T., Ben-Hamu, H., Nickel, M., and Le, M.
\newblock Flow matching for generative modeling.
\newblock In \emph{International Conference on Learning Representations}, 2022.

\bibitem[Liu et~al.(2022)Liu, Gong, and Liu]{iclr22_rect}
Liu, X., Gong, C., and Liu, Q.
\newblock Flow straight and fast: Learning to generate and transfer data with rectified flow.
\newblock \emph{arXiv preprint arXiv:2209.03003}, 2022.

\bibitem[Liu et~al.(2021)Liu, Lin, Cao, Hu, Wei, Zhang, Lin, and Guo]{SWIN-T}
Liu, Z., Lin, Y., Cao, Y., Hu, H., Wei, Y., Zhang, Z., Lin, S., and Guo, B.
\newblock Swin transformer: Hierarchical vision transformer using shifted windows.
\newblock In \emph{International Conference on Computer Vision}, pp.\  10012--10022, 2021.

\bibitem[Lu et~al.(2022)Lu, Zhou, Bao, Chen, Li, and Zhu]{dpm_solver}
Lu, C., Zhou, Y., Bao, F., Chen, J., Li, C., and Zhu, J.
\newblock Dpm-solver: A fast ode solver for diffusion probabilistic model sampling in around 10 steps.
\newblock In \emph{Neural Information Processing Systems}, New Orleans, LA, USA, Nov.-Dec. 2022. NIPS.

\bibitem[Meng et~al.(2022)Meng, Gao, Kingma, Ermon, Ho, and Salimans]{cvpr22_kd_guided}
Meng, C., Gao, R., Kingma, D.~P., Ermon, S., Ho, J., and Salimans, T.
\newblock On distillation of guided diffusion models.
\newblock \emph{arXiv preprint arXiv:2210.03142}, 2022.

\bibitem[Park \& Kim(2021)Park and Kim]{howvitwork}
Park, N. and Kim, S.
\newblock How do vision transformers work?
\newblock In \emph{International Conference on Learning Representations}, Virtual Event, May 2021. Openreview.net.

\bibitem[Russakovsky et~al.(2015)Russakovsky, Deng, Su, Krause, Satheesh, Ma, Huang, Karpathy, Khosla, Bernstein, et~al.]{ILSVRC15}
Russakovsky, O., Deng, J., Su, H., Krause, J., Satheesh, S., Ma, S., Huang, Z., Karpathy, A., Khosla, A., Bernstein, M., et~al.
\newblock Imagenet large scale visual recognition challenge.
\newblock \emph{International Journal of Computer Vision}, 115\penalty0 (3):\penalty0 211--252, 2015.

\bibitem[S \& J(1997)S and J]{LSTM}
S, H. and J, S.
\newblock Long short-term memory.
\newblock \emph{Neural Computation}, pp.\  1375--1780, 1997.

\bibitem[Sandler et~al.(2018)Sandler, Howard, Zhu, Zhmoginov, and Chen]{mobilenetv2}
Sandler, M., Howard, A.~G., Zhu, M., Zhmoginov, A., and Chen, L.
\newblock Mobilenetv2: Inverted residuals and linear bottlenecks.
\newblock In \emph{Computer Vision and Pattern Recognition}, pp.\  4510--4520, Salt Lake City, UT, USA, Jun. 2018. IEEE.

\bibitem[Shao et~al.(2023)Shao, Chen, Huang, Gong, Wang, and Wu]{TST}
Shao, S., Chen, H., Huang, Z., Gong, L., Wang, S., and Wu, X.
\newblock Teaching what you should teach: {A} data-based distillation method.
\newblock In \emph{International Joint Conference on Artificial Intelligence}, pp.\  1351--1359, Macao, SAR, China, Aug. 2023. ijcai.org.

\bibitem[Shen \& Xing(2022)Shen and Xing]{FastKD}
Shen, Z. and Xing, E.
\newblock A fast knowledge distillation framework for visual recognition.
\newblock In \emph{European Conference on Computer Vision}, pp.\  673--690. Springer, 2022.

\bibitem[Solodskikh et~al.(2023)Solodskikh, Kurbanov, Aydarkhanov, Zhelavskaya, Parfenov, Song, and Lefkimmiatis]{INN}
Solodskikh, K., Kurbanov, A., Aydarkhanov, R., Zhelavskaya, I., Parfenov, Y., Song, D., and Lefkimmiatis, S.
\newblock Integral neural networks.
\newblock In \emph{Computer Vision and Pattern Recognition}, pp.\  16113--16122, Vancouver, BC, Jun. 2023. {IEEE}.

\bibitem[Song \& Chai(2018)Song and Chai]{CL}
Song, G. and Chai, W.
\newblock Collaborative learning for deep neural networks.
\newblock In \emph{Neural Information Processing Systems}, pp.\  1837--1846, Montréal Canada, Dec. 2018. MIT Press.

\bibitem[Song et~al.(2023{\natexlab{a}})Song, Meng, and Ermon]{ddim}
Song, J., Meng, C., and Ermon, S.
\newblock Denoising diffusion implicit models.
\newblock In \emph{International Conference on Learning Representations}, kigali, rwanda, May. 2023{\natexlab{a}}. OpenReview.net.

\bibitem[Song \& Ermon(2019)Song and Ermon]{ncsn}
Song, Y. and Ermon, S.
\newblock Generative modeling by estimating gradients of the data distribution.
\newblock In \emph{Neural Information Processing Systems}, pp.\  11895--11907, Vancouver, BC, Canada, Dec. 2019.

\bibitem[Song et~al.(2023{\natexlab{b}})Song, Dhariwal, Chen, and Sutskever]{icml23_consistency}
Song, Y., Dhariwal, P., Chen, M., and Sutskever, I.
\newblock Consistency models.
\newblock \emph{arXiv preprint arXiv:2303.01469}, 2023{\natexlab{b}}.

\bibitem[Song et~al.(2023{\natexlab{c}})Song, Sohl-Dickstein, Kingma, Kumar, Ermon, and Poole]{sde}
Song, Y., Sohl-Dickstein, J., Kingma, D.~P., Kumar, A., Ermon, S., and Poole, B.
\newblock Score-based generative modeling through stochastic differential equations.
\newblock In \emph{International Conference on Learning Representations}, kigali, rwanda, May. 2023{\natexlab{c}}. OpenReview.net.

\bibitem[Szegedy et~al.(2015)Szegedy, Liu, Jia, Sermanet, Reed, Anguelov, Erhan, Vanhoucke, and Rabinovich]{Inceptionv1}
Szegedy, C., Liu, W., Jia, Y., Sermanet, P., Reed, S., Anguelov, D., Erhan, D., Vanhoucke, V., and Rabinovich, A.
\newblock Going deeper with convolutions.
\newblock In \emph{Computer Vision and Pattern Recognition}, Boston, Massachusetts, Jun. 2015. IEEE.

\bibitem[Tao et~al.(2022)Tao, Shan, Fei, Chen, and Xu]{DIST}
Tao, H., Shan, Y., Fei, W., Chen, Q., and Xu, C.
\newblock Knowledge distillation from a stronger teacher.
\newblock In \emph{Advances in Neural Information Processing Systems}, 2022.

\bibitem[Tian et~al.(2019)Tian, Krishnan, and Isola]{CRD}
Tian, Y., Krishnan, D., and Isola, P.
\newblock Contrastive representation distillation.
\newblock In \emph{International Conference on Learning Representations}, 2019.

\bibitem[Touvron et~al.(2021)Touvron, Cord, Douze, Massa, Sablayrolles, and J{\'e}gou]{DeiT}
Touvron, H., Cord, M., Douze, M., Massa, F., Sablayrolles, A., and J{\'e}gou, H.
\newblock Training data-efficient image transformers \& distillation through attention.
\newblock In \emph{International conference on machine learning}, pp.\  10347--10357. PMLR, 2021.

\bibitem[Tung \& Mori(2019)Tung and Mori]{SPKD}
Tung, F. and Mori, G.
\newblock Similarity-preserving knowledge distillation.
\newblock In \emph{International Conference on Computer Vision}, pp.\  1365--1374, 2019.

\bibitem[Wang et~al.(2022)Wang, Lohit, Jones, and Fu]{nips_kd_aug_statist}
Wang, H., Lohit, S., Jones, M.~N., and Fu, Y.
\newblock What makes a "good" data augmentation in knowledge distillation - a statistical perspective.
\newblock In \emph{Neural Information Processing Systems}, volume~35, pp.\  13456--13469, New Orleans, LA, USA, Dec. 2022. NIPS.

\bibitem[Wightman et~al.(2021)Wightman, Touvron, and Jegou]{timm}
Wightman, R., Touvron, H., and Jegou, H.
\newblock Resnet strikes back: An improved training procedure in timm.
\newblock In \emph{NeurIPS 2021 Workshop on ImageNet: Past, Present, and Future}, 2021.
\newblock URL \url{https://openreview.net/forum?id=NG6MJnVl6M5}.

\bibitem[Williams \& Zipser(1989)Williams and Zipser]{RNN}
Williams, R.~J. and Zipser, D.
\newblock A learning algorithm for continually running fully recurrent neural networks.
\newblock \emph{Neural Computation}, 1\penalty0 (2):\penalty0 270--280, 1989.

\bibitem[Xu et~al.(2020)Xu, Liu, Li, and Loy]{SSKD}
Xu, G., Liu, Z., Li, X., and Loy, C.~C.
\newblock Knowledge distillation meets self-supervision.
\newblock In Vedaldi, A., Bischof, H., Brox, T., and Frahm, J.-M. (eds.), \emph{European Conference on Computer Vision}, pp.\  588--604, Cham, 2020. Springer.

\bibitem[Yang et~al.(2022{\natexlab{a}})Yang, Li, Shao, Shi, Yuan, and Yuan]{Maskd}
Yang, Z., Li, Z., Shao, M., Shi, D., Yuan, Z., and Yuan, C.
\newblock Masked generative distillation.
\newblock In \emph{European Conference Computer Vision}, volume 13671, pp.\  53--69, Tel Aviv, Israel, Oct. 2022{\natexlab{a}}. Springer.

\bibitem[Yang et~al.(2022{\natexlab{b}})Yang, Li, Zeng, Li, Yuan, and Li]{ViTKD}
Yang, Z., Li, Z., Zeng, A., Li, Z., Yuan, C., and Li, Y.
\newblock Vitkd: Practical guidelines for vit feature knowledge distillation.
\newblock \emph{arXiv preprint arXiv:2209.02432}, 2022{\natexlab{b}}.

\bibitem[Yao et~al.(2024)Yao, Lu, Zhang, Zhang, Zhao, and Yu]{kdiffusion}
Yao, X., Lu, F., Zhang, Y., Zhang, X., Zhao, W., and Yu, B.
\newblock Progressively knowledge distillation via re-parameterizing diffusion reverse process.
\newblock In \emph{Association for the Advance of Artificial Intelligence}, 2024.

\bibitem[Zagoruyko \& Komodakis(2016{\natexlab{a}})Zagoruyko and Komodakis]{ATKD}
Zagoruyko, S. and Komodakis, N.
\newblock Paying more attention to attention: Improving the performance of convolutional neural networks via attention transfer.
\newblock In \emph{International Conference on Learning Representations}, 2016{\natexlab{a}}.

\bibitem[Zagoruyko \& Komodakis(2016{\natexlab{b}})Zagoruyko and Komodakis]{Zagoruyko2016WRN}
Zagoruyko, S. and Komodakis, N.
\newblock Wide residual networks.
\newblock In \emph{British Machine Vision Conference}, York, UK, Spet. 2016{\natexlab{b}}. {BMVA} Press.

\bibitem[Zhang et~al.(2022)Zhang, Malkin, Liu, Volokhova, Courville, and Bengio]{gflownet2}
Zhang, D., Malkin, N., Liu, Z., Volokhova, A., Courville, A.~C., and Bengio, Y.
\newblock Generative flow networks for discrete probabilistic modeling.
\newblock In \emph{International Conference on Machine Learning}, volume 162, pp.\  26412--26428, Baltimore, Maryland, {USA}, Jul. 2022. {PMLR}.

\bibitem[Zhang \& Ma(2020)Zhang and Ma]{FKD}
Zhang, L. and Ma, K.
\newblock Improve object detection with feature-based knowledge distillation: Towards accurate and efficient detectors.
\newblock In \emph{International Conference on Learning Representations}, 2020.

\bibitem[Zhang et~al.(2018)Zhang, Zhou, Lin, and Sun]{shufflenet}
Zhang, X., Zhou, X., Lin, M., and Sun, J.
\newblock Shufflenet: An extremely efficient convolutional neural network for mobile devices.
\newblock In \emph{Computer Vision and Pattern Recognition}, pp.\  6848--6856, 2018.

\bibitem[Zhao et~al.(2022)Zhao, Cui, Song, Qiu, and Liang]{DKD}
Zhao, B., Cui, Q., Song, R., Qiu, Y., and Liang, J.
\newblock Decoupled knowledge distillation.
\newblock In \emph{Computer Vision and Pattern Recognition}, pp.\  11953--11962, June 2022.

\bibitem[Zhong et~al.(2020)Zhong, Zheng, Kang, Li, and Yang]{RandomEasing}
Zhong, Z., Zheng, L., Kang, G., Li, S., and Yang, Y.
\newblock Random erasing data augmentation.
\newblock In \emph{Association for the Advance of Artificial Intelligence}, volume~34, pp.\  13001--13008, 2020.

\bibitem[Zong et~al.(2023)Zong, Qiu, Ma, Yang, Liu, Hou, Yi, and Ouyang]{DPK}
Zong, M., Qiu, Z., Ma, X., Yang, K., Liu, C., Hou, J., Yi, S., and Ouyang, W.
\newblock Better teacher better student: Dynamic prior knowledge for knowledge distillation.
\newblock In \emph{International Conference on Learning Representations}, Kigali, Rwanda, May 2023. OpenReview.net.
\newblock URL \url{https://openreview.net/pdf?id=M0\_sUuEyHs}.

\end{thebibliography}
\bibliographystyle{icml2023}

\newpage
\appendix
\onecolumn
{\centering \textbf{APPENDIX}}
\begin{itemize}
	\item[] {\color{red}Appendix} \ref{apd:pseudocode}: Pseudo Code of FM-KT.
	\item[] {\color{red}Appendix} \ref{apd:training_paradiam}: Theoretical Guarantees of FM-KT.
    \item[] {\color{red}Appendix} \ref{apd:link_ensemble_with_flow}: Link FM-KT to Ensemble.
    \item[] {\color{red}Appendix} \ref{apd:pair_decoupling}: Pair Decoupling.
    \item[] {\color{red}Appendix} \ref{apd:add_ablation}: Additional Ablation Experiment.
    \item[] {\color{red}Appendix} \ref{apd:cnr}: Additional Related Work on Continuous Network Representations.
    \item[] {\color{red}Appendix} \ref{apd:obj}: Additional Object Detection Comparison.
    \item[] {\color{red}Appendix} \ref{sec:stronger}: Stronger Strategies and Stronger Teacher Comparison.
    \item[] {\color{red}Appendix} \ref{sec:deit}: Vision Transformer Comparison.
    \item[] {\color{red}Appendix} \ref{apd:visualization}: Visualization of Sampling Trajectory.
    \item[] {\color{red}Appendix} \ref{apd:implementation_detail}: Implementation Detail.
    \begin{enumerate}
        \item[] \ref{apd:implementation_detail:training_strategies}: Training Strategies.
        \item[] \ref{:apd:implementation_detail:loss_function}: Loss Function and Meta-encoder.
    \end{enumerate}
    \item[] {\color{red}Appendix} \ref{apd:add_training_and_inference_cc_discussion}: Additional Training and Inference Computational Cost Discussion.
    \item[] {\color{red}Appendix} \ref{apd:best_meta_encoder}: Best Meta-encoder Choice on ImageNet-1k.
    \item[] {\color{red}Appendix} \ref{apd:arch_sensitive}: Architecture-Sensitive Experiments between FM-KT and DiffKD.
    \item[] {\color{red}Appendix} \ref{apd:unified}: Unify VP SDE, VE SDE and Rectified flow in FM-KT.
    \item[] {\color{red}Appendix} \ref{apd:limitation}: Limitation.
\end{itemize}

\section{Pseudo Code of FM-KT}
\label{apd:pseudocode}
For ease of understanding, we show the pseudo code of FM-KT in the Offline KD scenario. The implementation of FM-KT$^\Theta$ and OFM-KT only needs to modify the optimization objective as described in our main paper.
\begin{algorithm}[h!]
    \caption{Pseudo code of FM-KT in a PyTorch-like style.}
    \label{algo:fm_kd}
    \tiny
    \begin{alltt}
import torch
import torch.nn as nn
import torch.nn.functional as F

\color{Black}class FlowMatchingModule(nn.Module):
    def __init__(self,...):
        super().__init__()
        self.meta_encoder:nn.Module = (...)
        self.metric_based_loss_function:nn.Module = (...)
        self.time_embed:nn.Module = nn.Linear(...)
        self.training_sampling:int = (...) \color{ForestGreen} # the number of sampling steps during training
        \color{Black}self.shape_transformation_function:nn.Module = (...)
        self.dirac_ratio:float = (...) \color{ForestGreen} # hyperparameter \(\beta\sb{d}\), which belongs to [0,1]
        \color{Black}self.weight:float = (...)
\end{alltt}
\vspace{-5pt}
\begin{alltt}
\color{Black}
    def forward(self, s_f, t_f=None, target=None, inference_sampling=1):
        \color{ForestGreen}# s_f: the feature/logit of the student
        # t_f: the feature/logit of the teacher
        # target: the logit-based ground truth label, only used for logit-based distillation
        # inference_sampling: the number of sampling steps during inference
        
        \color{Black}all_p_t_f = []
        \color{Black}if self.training:
            \color{ForestGreen}# Shuffle one-to-one teacher-student feature/logit pair
            \color{Black}if t_f is not None:
                l = int(self.dirac_ratio * t_f.shape[0])
                t_f[l:][torch.randperm(t_f.shape[0] - l)] = t_f[l:].clone()
            loss, x = 0., s_f
            indices = reversed(range(1, self.training_sampling + 1))
            \color{ForestGreen}# Calculate the FM-KT loss
            \color{Black}for i in indices:
                t = torch.ones(s_f.shape[0]) * i / self.training_sampling
                embed_t = self.time_embed(t)
                embed_x = x + embed_t
                velocity = self.meta_encoder(embed_x)
                x = x - velocity / self.training_sampling
                p_t_f = self.shape_transformation_function(s_f - velocity)
                all_p_t_f.append(p_t_f)
                loss += self.metric_based_loss_function(p_t_f, t_f)
                if target is not None:
                    loss += F.cross_entropy(p_t_f, target)
            loss *= (self.weight / self.training_sampling)
            return loss, torch.stack(all_p_t_f, 0).mean(0)
        else:
            x = s_f
            indices = reversed(range(1, inference_sampling + 1))
            \color{Black}for i in indices:
                t = torch.ones(s_f.shape[0]) * i / inference_sampling
                embed_t = self.time_embed(t)
                embed_x = x + embed_t
                velocity = self.meta_encoder(embed_x)
                x = x - velocity / inference_sampling
                all_p_t_f.append(self.shape_transformation_function(s_f - velocity))
            return 0., torch.stack(all_p_t_f, 0).mean(0)
\end{alltt}
\end{algorithm}

\section{Theoretical Guarantees of FM-KT}
\label{apd:training_paradiam}
FM-KT proposes a novel serial training paradigm in order to avoid ``cheating'' by accessing $X^T$ during training:
\begin{equation}
\small
\begin{aligned}
&  \mathcal{L}_\textrm{FM-KT} = \mathbb{E}_{(X^S,X^T)} \frac{1}{N}\sum_{i=1}^{N} ||\mathcal{T}((\nabla_t\alpha_t Z_1 - g_{v_\theta}(Z_{1-i/N},1-i/N))/-\nabla_t\sigma_t)- X^T||_2^2,\\
\end{aligned}
\label{eq:theoretical_guarantee_1}
\end{equation}
where $Z_1=\alpha_1X^S$. Here we assume the loss function is $\ell_2$-norm. Broadly speaking, the loss function used by FM-KT only needs to ensure that it can achieve the effect of minimizing the difference in distributions similar to Kullback-Leibler Divergence. When $i\geq1$, $Z_{1-i/N} = Z_{1-(i-1)/N} - g_{v_\theta}(Z_{1-(i-1)/N},1-(i-1)/N)/N$. $\mathcal{T}(\cdot)$ is used for shape alignment to ensure the calculation of $\ell_2$-norm. Let us define $q(\cdot|\cdot)$ as the predefined conditional probability, and $p_{v_\theta}(\cdot|\cdot)$ as the predicted conditional probability. We know the training objective of the classical diffusion probability model can be performed by minimizing the upper bound on negative log-likelihood:
\begin{equation}
\small
\begin{aligned}
& -\log p_{v_\theta}(Z_0) \leq \mathbb{E}_{q(Z_{1/N:1}|Z_0)}\left[\log\frac{q(Z_{1}|Z_0)}{p_{v_\theta}(Z_1)p_{v_\theta}(Z_0|Z_{1/N})}+\sum_{i=1}^N\log \frac{q(Z_{(i-1)/N}|Z_{i/N},Z_0)}{p_{v_\theta}(Z_{(i-1)/N}|Z_{i/N})}\right].\\
\end{aligned}
\label{eq:ddpm_origin}
\end{equation}
We can rewritten it as
\begin{equation}
\small
\begin{aligned}
& -\log p_{v_\theta}(Z_0) \leq \mathbb{E}_{q(Z_{1/N:1}|Z_0)}\left[\log\frac{q(Z_{1}|Z_0)}{p_{v_\theta}(Z_1)p_{v_\theta}(Z_0|Z_{1/N})}+\sum_{i=1}^N\log \frac{q(Z_{(i-1)/N}|Z_{i/N},Z_0)}{p_{v_\theta}(Z_{(i-1)/N}|Z_{i/N})}\right]\\
& = \mathbb{E}_{q(Z_{1/N:1}|Z_0)}\left[\log\frac{q(Z_{1}|Z_0)}{p_{v_\theta}(Z_1)p_{v_\theta}(Z_0|Z_{1/N})}\right]+\sum_{i=1}^N \mathbb{E}_{q(Z_{i/N}|Z_0)}\mathbb{E}_{q(Z_{(i-1)/N}|Z_{i/N},Z_0)}\left[\log \frac{q(Z_{(i-1)/N}|Z_{i/N},Z_0)}{p_{v_\theta}(Z_{(i-1)/N}|Z_{i/N})}\right] \\
& = \mathbb{E}_{q(Z_{1/N:1}|Z_0)}\left[\log\frac{q(Z_{1}|Z_0)}{p_{v_\theta}(Z_1)p_{v_\theta}(Z_0|Z_{1/N})}\right]+\sum_{i=1}^N \mathbb{E}_{\hat{Z}_{i/N} \sim \int p_{v_\theta}(Z_{i/N}|Z_1)q(Z_1|Z_0)dZ_1} \\
&\left[\KL(q(Z_{(i-1)/N}|Z_{i/N},Z_0)||p_{v_\theta}(Z_{(i-1)/N}|\hat{Z}_{i/N}))\right],\quad s.t.\quad \textrm{Law}(Z_{i/N}) \stackrel{\sim}{=}\textrm{Law}(\hat{Z}_{i/N}) \\
& \approx \mathbb{E}_{q(Z_{1/N:1}|Z_0)}\left[\log\frac{q(Z_{1}|Z_0)}{p_{v_\theta}(Z_1)p_{v_\theta}(Z_0|Z_{1/N})}\right]+\sum_{i=1}^N \mathbb{E}_{\hat{Z}_{i/N} \sim \int p_{v_\theta}(Z_{i/N}|Z_1)q(Z_1|Z_0)dZ_1} \\
&\left[||q(Z_{(i-1)/N}|Z_{i/N},Z_0)-p_{v_\theta}(Z_{(i-1)/N}|\hat{Z}_{i/N})||_2^2\right],\quad s.t.\quad \textrm{Law}(Z_{i/N}) \stackrel{\sim}{=}\textrm{Law}(\hat{Z}_{i/N}).
\end{aligned}
\label{eq:FM-KT_origin_1}
\end{equation}
For $i\!\geq\!1$, if $\textrm{Law}(Z_{i/N})\stackrel{\sim}{=} \textrm{Law}(\hat{Z}_{i/N})$ is guaranteed, then $ \textrm{Law}(Z_{(i-1)/N})\stackrel{\sim}{=} \textrm{Law}(\hat{Z}_{(i-1)/N})$ can also be guaranteed by optimizing $\mathbb{E}_{\hat{Z}_{i/N},Z_1,Z_0}\KL(q(Z_{(i-1)/N}|Z_{i/N},Z_0)||p_{v_\theta}(Z_{(i-1)/N}|\hat{Z}_{i/N}))$ in Eq.~\ref{eq:FM-KT_origin_1}. Based on the prior condition $\textrm{Law}(Z_{1})\stackrel{\sim}{=} \textrm{Law}(\hat{Z}_{1})$, we can deduce $\{\textrm{Law}(Z_{i/N})\stackrel{\sim}{=} \textrm{Law}(\hat{Z}_{i/N})\}_{i=0}^{N-1}$ sequentially by recursive method.

Note that this derivation via Bayes' Theorem satisfies almost all noise schedules, such as Rectified flow~\citep{iclr22_rect,CNF}, VP ODE~\citep{sde,stochastic_interpolant} and VE ODE~\citep{sde,stochastic_interpolant}. More details can be found in Appendix~\ref{apd:unified}. In fact, the upper bound in Eq.~\ref{eq:FM-KT_origin_1} is precisely the optimization objective of FM-KT. The main difference between Eq.~\ref{eq:ddpm_origin} and Eq.~\ref{eq:FM-KT_origin_1} is that $\hat{Z}_{i/N}$ replaces ${Z}_{i/N}$, and $\hat{Z}_{i/N}$ is obtained from the reverse sampling process. In this manner, the trajectory $\{\hat{Z}_{i/N}\}_i$ is obtained through the estimator $g_{v_\theta}(\cdot)$ and no longer contains the teacher knowledge $X^T$, thereby enabling the distillation process to proceed normally.

\section{Link FM-KT to Ensemble}
\label{apd:link_ensemble_with_flow}
Ensemble is a method that trains multiple models, aggregates their outputs through voting, and produces a final prediction. In this section, we prove theoretically that FM-KT is essentially a unique implicit ensemble method under the assumption that its noise schedule is set as Rectified flow.

First, we define ODE in FM-KT as $X^S - X^T =\frac{dX_t}{dt}$ (for the convenience of derivation, this definition is slightly different from that in the main paper), so we need to fit $||\frac{dX_t}{dt} - g_{v_\theta}(X_t,t)||_2^2$, where $X_t =tX^S+(1-t)X^T$, $t\sim \mathcal{U}[0,1]$. In inference, this ODE solver defaults to Euler's method in FM-KT, and the sampling must be discrete with $N$ steps because fitting continuous time steps $t$ consumes extensive computational costs. When the meta-encoder $g_{v_\theta}(\cdot)$ is at the optimal solution, we assume that its error from the true value can be expressed as a function of $x_t$ and $t$, and that this function is at 1-Lipschitz. 

Thus, we can define $\mathcal{H}(t) = \operatorname*{arg\,sup}_{X_t} \{||\frac{dX_t}{dt}- g_{v_\theta}(X_t,t)||_2^2\}$, then the truncation error $\mathcal{K}(t)$ can be defined as $\left[\frac{dX_t}{dt} -  g_{v_\theta}(\mathcal{H}(t),t)\right]$, which is also at 1-Lipschitz under the assumption $\mathcal{H}(t)$ is at 1-Lipschitz. After that, we also need to define the step number of sampling in inference. We set it as $K$, so $dt$ is $1/K$. Based on the aforementioned notations, we can analyse the truncation error by the recursive method.

For the sake of derivation convenience, we define $\{Z_t\}_t$ as the sampled trajectory in inference to distinguish it from $\{X_t\}_t$ in training. Thus, a step in sampling can be described as $Z_{1-(i+1)/K}\!=\!Z_{1-i/K}-g_{v_\theta}(Z_{1-{i/K}},1-i/K)dt$, and $Z_{1-i/K} = X_{1-i/K}+\mathcal{E}(Z_{1-i/K})$, where $\mathcal{E}(Z_{1-i/K})$ refers to the truncation error accumulated to a intermediate sample $Z_{1-i/K}$ in the sampling process. Note that $\mathcal{K}(t)$ and $\mathcal{E}(Z_{1-i/K})$ are not results of the norm, and therefore $\forall t$ and $\forall Z_{1-i/K}$, this derivation does not need to satisfy that $\mathcal{K}(t)\geq 0$ and $\mathcal{E}(Z_{1-i/K}) \geq 0$. This approach avoids the accumulation of the truncation error due to $\ell_2$-norm $\geq$ 0. We can derive the sample $Z_{1-(i-1)/K}$ in the next step by the derivation:
\begin{equation}
\small
\begin{aligned}
Z_{1-(i+1)/K} & = Z_{1-i/K} - (1/K)g_{v_\theta}(Z_{1-i/K},1-i/K) \\
 & = Z_{1-i/K} - (1/K)g_{v_\theta}(X_{1-i/K}+\mathcal{E}(Z_{1-i/K}),1-i/K) \\
& = X_{1-i/K} + \mathcal{E}(Z_{1-i/K}) - (1/K)g_{v_\theta}(X_{1-i/K}+\mathcal{E}(Z_{1-i/K}),1-i/K) \\
& \approx X_{1-i/K} + \mathcal{E}(Z_{1-i/K})-(1/K)\left[g_{v_\theta}(X_{1-i/K},1-i/K)+ \mathcal{E}(Z_{1-i/K})\nabla_{X_t}g_{v_\theta}(X_{1-i/K},1-i/K)\right], \\
& = X_{1-i/K} + \mathcal{E}(Z_{1-i/K}) -(1/K)\left[g_{v_\theta}(X_{1-i/K},1-i/K)+ \mathcal{E}(Z_{1-i/K})\psi(1-i/K)\right], \\
\end{aligned}
\label{eq:flow_recursive_1}
\end{equation}
where $\psi(t) = \nabla_{X_t}g_{v_\theta}(X_t,t)$. Then, Eq.~\ref{eq:flow_recursive_1} can continue to be derived as
\begin{equation}
\small
\begin{aligned}
Z_{1-(i+1)/K} & \approx X_{1-(i+1)/K}+\mathcal{E}(Z_{1-i/K}) + (1/K)\mathcal{K}(1-i/K)-(1/K)\mathcal{E}(Z_{1-i/K})\psi(1-i/K) \\
& = X_{1-(i+1)/K} + \mathcal{E}(Z_{1-i/K})[1-(1/K)\psi(1-i/K)]+(1/K) \mathcal{K}(1-i/K) \\
Z_{1-(i+1)/K} - X_{1-(i+1)/K} & =  \mathcal{E}(Z_{1-i/K})[1-(1/K)\psi(1-i/K)]+(1/K) \mathcal{K}(1-i/K). \\
\end{aligned}
\label{eq:flow_recursive_2}
\end{equation}
Thus, $\mathcal{E}(Z_{1-(i+1)/K}) = \mathcal{E}(Z_{1-i/K})[1-(1/K)\psi(1-i/K)] + (1/K) \mathcal{K}(1-i/K)$. After that, the recursive method leads us to the following conclusions:
\begin{equation}
\small
\begin{aligned}
    & \mathcal{E}(Z_{1-1/K}) = (1/K)\mathcal{K}(1) \\
& \mathcal{E}(Z_{1-2/K}) = \mathcal{E}(Z_{1-1/K})(1-(1/K)\psi(1-1/K))+(1/K)\mathcal{K}(1-1/K) \\
& \vdots \\
& \mathcal{E}(Z_0) = (1/K)\left[\sum_{i=0}^{K-1}\mathcal{K}(1-i/K)\right] + (1/K^2)\left[\sum_{j=1}^{K-1}\left[\psi(1-j/K)\left(\sum_{i=0}^{j-1}\mathcal{K}(1-i/K)\right)\right]\right] + \mathcal{O}(1/K^3).\\
\end{aligned}
\label{eq:flow_recursive_3}
\end{equation}
Looking at the first term, we can see that the truncation error comes from summing $\mathcal{K}(\cdot)$ over all time points. When treating the error sampling as Monte Carlo sampling, with a sufficient number of samplings $K$, it becomes possible for FM-KT to approximate ensemble methods and thus estimate the ground truth effectively.

\section{Pair Decoupling}
\label{apd:pair_decoupling}
In this section, we present pair decoupling (PD), a straightforward yet effective technique for enhancing performance in the feature-based distillation scenario of image classification using FM-KT. This method involves shuffling a subset of samples in a batch to achieve regularization, thereby preventing overfitting of the teacher's refined low-level hierarchical features. Let $B$, $C$, $H$ and $W$ denote the batch size, the number of channels, the height of the feature map, and the width of the feature map, respectively. Given the teacher's feature map $X^T\in \mathbb{R}^{B\times C\times H \times W}$ from a specific layer, PD is applied prior to all FM-KT related calculations. Implementing PD involves defining a hyperparameter, the dirac ratio $\beta_d$, and perturbing $B\!-\!\left\lfloor {\beta_d}B \right\rfloor$ samples in the batch. The Pytorch code for this is provided in Appendix~\ref{apd:pseudocode}. Specifically, PD selects the random $B\!-\!\left\lfloor {\beta_d}B \right\rfloor$ samples $X^T\left[0:B\!-\!\left\lfloor {\beta_d}B \right\rfloor\right]$ in a batch and then shuffles them: $$X^T\left[0:B\!-\!\left\lfloor {\beta_d}B\right\rfloor\right]= \textbf{\textrm{shuffle}}\left(X^T\left[0:B\!-\!\left\lfloor {\beta_d}B \right\rfloor\right]\right).$$

We refer to the hyperparameter $\beta_d$ as ``dirac ratio'' because, following the PD operation, $\left\lfloor {\beta_d}B \right\rfloor$ samples are used to compute $\mathcal{L}_\textrm{FM-KT}$. Here, $X^T$ and $X^S$ are treated as Dirac distributions with the objective of achieving one-to-one matching. Conversely, the remaining $B - \left\lfloor {\beta_d}B \right\rfloor$ samples are utilized in the computation of $\mathcal{L}_\textrm{FM-KT}$, where $X^T$ and $X^S$ are considered as non-Dirac distributions, targeting many-to-many matching.

Due to the specificity of the feature-based distillation scenario for image classification, PD is specifically designed to avoid over-matching the refined low-level feature thus improving the final performance of the student. Meanwhile, our experiments in Appendix~\ref{apd:add_ablation} empirically demonstrate that PD is effective only in the feature-based distillation scenario of image classification, whereas in other scenarios it rather degrades the performance. This is because matching the teacher's feature/logit at a fine-grained level is closely related to the final performance of the student in the logit-based distillation scenario for image classification as well as the feature-based distillation scenario for object detection. In other words, in the feature-based distillation scenario for image classification, it does not imply that improving similarity between the student's low-level feature and the teacher's low-level feature will result in the greater classification accuracy of the student.

\section{Additional Ablation Experiment}
\label{apd:add_ablation}
Here, we experimentally substantiate some empirical findings on the topics of normalization layer selection in the meta-encoder, where stages used for distillation in the feature-based scenario, and ideal configuration of dirac ratio $\beta_d$ in different scenarios.

\begin{wraptable}{r}{5.8cm}
\tiny
\vspace{-14pt}
\renewcommand\arraystretch{0.85}
\setlength\tabcolsep{0.8mm}
\begin{center}
\begin{tabular}{l|cc}
\hline
\makecell{Normalization\\type} & GroupNorm & BatchNorm \\
\hline
\noalign{\smallskip}
WRN-40-2 (T) & 75.61 & 75.61 \\
WRN-40-2 (S+Baseline) & - & 73.26 \\
WRN-40-2 (S+DIST) & - & 75.29 \\
WRN-16-2 (S+FM-KT $K$=1) & 75.58 & 1.00 \\
WRN-16-2 (S+FM-KT $K$=2) & 75.85 & 1.00 \\
WRN-16-2 (S+FM-KT $K$=4) & 75.87 & 1.10 \\
WRN-16-2 (S+FM-KT $K$=8) & 75.87 & 1.43 \\\hline
\end{tabular}
 \end{center}
 \vspace{-0.1in}
\caption{Experiments were conducted on the different normalization type of FM-KT on CIFAR-100. Note that in this table all the architecture of the meta-encoder and the form of the loss function are set as CNN and DIST, respectively.}
\label{tab:add_normal_type}
\vspace{-10pt}
\end{wraptable}Ablation experiments on various normalization operations reveal instability in the FM-KT training paradigm when using BatchNorm. As observed in Table~\ref{tab:add_normal_type}, the accuracy achieved with BatchNorm as the normalization layer is approximately 1\%, even when the training of the student remains stable (\textit{i.e.}, the loss is not NAN). This indicates that using BatchNorm in FM-KT introduces instability by computing the mean and variance of inputs at different time points during inference. It is essential to note that although BatchNorm is applied in DiffKD, this choice is justified as the student converges effectively with a sufficiently high number of layers in the Diffusion Model (\textit{i.e.} meta-encoder) mentioned in their work. Similar results are obtained in our studies by replacing the meta-encoder in FM-KT with Diffusion Model in DiffKD.

\begin{figure}[!h]
\begin{center}
\includegraphics[width=0.95\linewidth]{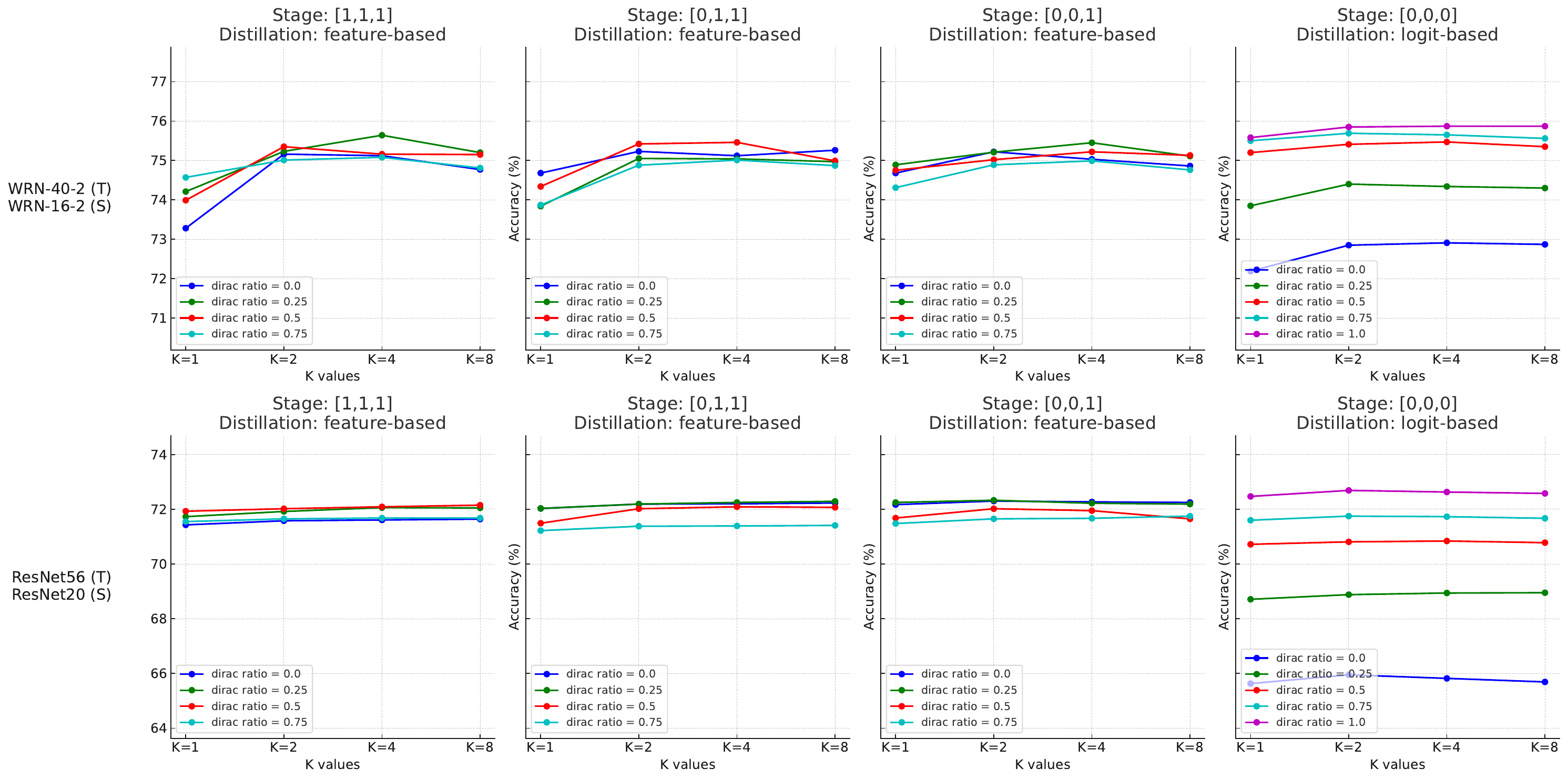}
\vspace{-10pt}
\end{center}
   \caption{Experiments were conducted on the various hyperparameter $\beta_d$ (dirac ratio) and different distillation stages of FM-KT on CIFAR-100. Note that in this figure all the architecture of the meta-encoder and the form of the loss function are set as CNN and DIST, respectively. The choice of the last three stages was fixed for distillation analysis. In the figure, the notation ``[n3, n2, n1]'' indicates that n\bb{x}=1 signifies the use of that the \bb{x}th last stage for distillation. For example, ``[0, 1, 1]'' signifies the utilization of the 1st and 2nd last stages for distillation.}
\label{fig:add_dirac_ratio}
\vspace{-10pt}
\end{figure}

In Fig.~\ref{fig:add_dirac_ratio}, we investigated the optimal stages for distillation in the feature-based scenario and the ideal configuration for the dirac ratio $\beta_d$. As there are no specific distillation stages for the logit-based distillation, we designated it as ``[0, 0, 0]'' for clarity. Our observations indicate that in the feature-based distillation scenario, the distillation stage does not significantly affect the final outcomes. Meanwhile, the configuration ``[1, 1, 1]'' often underperforms compared with ``[0, 1, 1]'' and ``[0, 0, 1]''. This observation aligns with the conclusions drawn from most of the prior feature-based distillation studies~\citep{SPKD,ATKD,DPK}. Moreover, for different values of $\beta_d$, the settings $\beta_d$=0.25 typically yields the best result in feature-based distillation, while $\beta_d$=1.0 excels in the logit-based distillation. This implies that the PD technique is more effective in the feature-based distillation context for image classification, and less so in the logit-based distillation.

It is worth noting that our experiments on PD in object detection revealed that $\beta_d$=1.0 and $\beta_d$=0.75 yield comparable performance, whereas a decrease in $\beta_d$ results in diminished performance. Based on these findings, we recommend using $\beta_d$=0.25 as the default in the feature-based distillation scenario for image classification and $\beta_d$=1.0 in other contexts.
\setlength{\tabcolsep}{4pt}
\begin{table*}[!t]
\tiny
\renewcommand\arraystretch{0.9}
\begin{center}
\resizebox{0.7\textwidth}{!}{%
\begin{tabular}{l|clccccc}
\hline
Method & Schedule & mAP & AP$_\textrm{50}$ & AP$_\textrm{75}$ & AP$_\textrm{S}$ & AP$_\textrm{M}$ & AP$_\textrm{L}$  \\
\hline
Mask RCNN-Swin (T) & 3$\times$+ms & 48.2 &  69.8 & 52.8 & 32.1 & 51.8 & 62.7 \\
\baseline{Retina-Res50 (S)} & \baseline{2$\times$} & \baseline{37.4} & \baseline{56.7} & \baseline{39.6} & \baseline{20.0} & \baseline{40.7} & \baseline{49.7} \\
PKD  & 2$\times$ & 41.3 (+3.9) & 60.5 & \textbf{44.1} & \textbf{23.0} & 45.3 & \textbf{55.9} \\
FM-KT ($K$=1) & 2$\times$ & \textbf{41.4 (+4.0)} & \textbf{60.6} & 44.0 & 22.5 & \textbf{45.6} & 55.7 \\
FM-KT ($K$=4) & 2$\times$ & \textbf{41.4 (+4.0)} & \textbf{60.6} & \textbf{44.1} & 22.5 & \textbf{45.6} & 55.7 \\
\hline
\noalign{\smallskip}
FasterRCNN-Res101 (T) & 2$\times$ & 39.8 & 60.1 & 43.3 & 22.5 & 43.6 & 52.8 \\
\baseline{FasterRCNN-Res50 (S)}  & \baseline{2$\times$} & \baseline{38.4} & \baseline{59.0} & \baseline{42.0} & \baseline{21.5} & \baseline{42.1} & \baseline{50.3} \\
GID & 2$\times$ & 40.2 (+1.8) & 60.7 & 43.8 & 22.7 & 44.0 & 53.2 \\
FRS & 2$\times$ & 40.4 (+2.0) & \textbf{60.8} & 44.0 & \textbf{23.2} & 44.4 & 53.1 \\
FGD & 2$\times$ & 40.4 (+2.0) & 60.7 & \textbf{44.3} & 22.8 & 44.5 & \textbf{53.5} \\
PKD & 2$\times$ & 40.3 (+1.9) & \textbf{60.8} & 44.0 & 22.9 & 44.5 & 53.1 \\
FM-KT ($K$=1) & 2$\times$ & 40.4 (+2.0) & 60.7 & 44.1 & 22.9 & \textbf{44.8} & 52.8 \\
FM-KT ($K$=4) & 2$\times$ & \textbf{40.5 (+2.1)} & 60.7 & 44.2 & 22.9 & \textbf{44.8} & 52.9 \\
\hline
\noalign{\smallskip}
FCOS-Res101 (T) & 2$\times$+ms & 41.2 & 60.4 & 44.2 & 24.7 & 45.3 & 52.7 \\
\baseline{Retina-Res50 (S)} & \baseline{1$\times$} & \baseline{37.4} & \baseline{56.7} & \baseline{39.6} & \baseline{20.0} & \baseline{40.7} & \baseline{49.7} \\
PKD  & 1$\times$ & 40.3 (+2.9) & 59.6 & 43.0 & 22.2 & 44.9 & \textbf{53.7} \\
FM-KT ($K$=1) & 1$\times$ & \textbf{40.5 (+3.1)} & \textbf{59.9} & \textbf{43.6} & \textbf{22.5} & \textbf{45.0} & 53.5 \\
FM-KT ($K$=4) & 1$\times$ & \textbf{40.5 (+3.1)} & 59.8 & \textbf{43.6} & \textbf{22.5} & \textbf{45.0} & \textbf{53.7} \\
\hline
\end{tabular}}
\vspace{-10pt}
\end{center}
\caption{Results of FM-KT with different detection frameworks on MS-COCO. ``T'' and ``S'' mean the ``teacher'' and ``student'' detector, respectively.}
\label{tab:ms_coco_results}
\vspace{-10pt}
\end{table*}
\section{Additional Related Work on Continuous Network Representations}
\label{apd:cnr}
With the development of deep learning, there are a number of architectures belonging to continuous network representation, such as RNN~\citep{RNN}, LSTM~\citep{LSTM}, Neural ODE~\citep{NODE}, GflowNet family~\citep{gflownet1,gflownet2}, diffusion model family~\citep{sde,nips22_design,ddpm_begin,ddim}, INN~\citep{INN}, DiffKD~\citep{DiffKD} and KDiffusion~\citep{kdiffusion}. Here, we mainly emphasize the similarities and differences between our proposed FM-KT and these methods. In this way, we show the novelty of the FM-KT design and its advantages in application:
\begin{itemize}
\item The application scenario of FM-KT is different from RNN, LSTM, Neural ODE, GflowNet family, diffusion model family and INN. Of these, only FM-KT, DiffKD and KDiffusion are applied to knowledge distillation in the form of continuous network representations.
\item RNN, LSTM, Neural ODE, GflowNet family, diffusion model family, INN, and FM-KT have a meta-encoder shared parameters. However, the difference is that the forward process (meaning the backward process in diffusion model family and FM-KT) in RNN, LSTM, and GflowNet family is unknown. Unlike Neural ODE, diffusion model family and FM-KT, there exists a human-designed sampling process (\textit{a.k.a} predefined forward processes), which makes it impossible to use numerical integration to trade-off performance and efficiency. 
\item INN primarily enables the continuous representation of convolutional operations (\textit{i.e.} convolutional kernels), not the entire network. In contrast, the Neural ODE, diffusion model family, continuously represents the entire network. 
\item The primary distinction between Neural ODE/FM-KT and the diffusion model family lies in their training paradigms. The diffusion model family is trained using unpaired samples, aiming to capture the entire data distribution. In contrast, Neural ODE/FM-KT utilizes paired samples, focusing on learning the Dirac distribution of the output.
\item The biggest difference between FM-KT and Neural ODE is that FM-KT has a deterministic a priori forward process to model the optimization objective of the intermediate points (\textit{i.e.} $\{Z_{1-i/N}\}_i$), which ensures the stability of training. Neural ODE has no such a priori forward process, and simply expects the network itself to learn a continuous representation from input to output.
\end{itemize}
\section{Object Detection Comparison}
\label{apd:obj}
The experimental results of object detection are presented in Table~\ref{tab:ms_coco_results}, where Mask RCNN-Swin-RetinaNet-Res50 pair represents the case of being distilled from a strong teacher, FasterRCNN-Res101-FasterRCNN-Res50 pair represents the homogeneous teacher-student pair, and FCOS-Res101-Retina-Res50 pair represents the heterogeneous teacher-student pair. We observe that FM-KT, which applies PKD as its loss function, shows improvement to some extent compared to the baseline FKD and achieves state-of-the-art performance across all teacher-student pairs. Note that knowledge transfer in object detection is facilitated by the high similarity between the feature maps of the student and the teacher. Consequently, the student's mAP remains consistent for both $K$=$4$ and $K$=$8$, so we do not present results for $K$=$8$.
\section{Stronger Strategies and Stronger Teacher Comparison}
\label{sec:stronger}
In recent years, with the advancement of deep learning, stronger training strategies and higher-quality foundational models have emerged. As a result, traditional distillation methods are no longer sufficient for capturing a superior student. In this context, we utilize the ResNet50 (with an accuracy of 80.1\%) from TIMM~\citep{timm} training as a stronger teacher to distill the ResNet18. Simultaneously, we adopt some stronger strategies: the learning rate begins at 5e-4, the chosen optimizer is AdamW, the batch size is set as 1024, the number of training epochs is set as 350, the learning rate warms up over 3 epochs, and then decays at a rate of 0.9874 per epoch. For data augmentation, we employ a combination of RandomCrop, RandomClip, RandAugment~\citep{RandAugment}, and RandomErasing~\citep{RandomEasing}. It's important to note that the loss function and the meta-encoder in FM-KT remain consistent with the main paper, being DIST and Swin-Transformer, respectively. Finally, the experimental results on ImageNet-1k are presented in Table~\ref{tab:stronger_results}.
\setlength{\tabcolsep}{4pt}
\begin{table}[!h]
\footnotesize
\renewcommand\arraystretch{0.9}
\begin{center}
\resizebox{1.0\textwidth}{!}{%
\begin{tabular}{lcccccc}
\hline
\noalign{\smallskip}
& Teacher (ResNet50) & \baseline{DIST} & FM-KT ($K$=1) & FM-KT ($K$=2) & FM-KT ($K$=4) & FM-KT ($K$=8) \\
\hline
\noalign{\smallskip}
Top-1 Acc. & 80.12\% & \baseline{72.89\%} & 72.61\% & 73.11\% & 73.59\% & 73.71\% \\
\hline
\end{tabular}}
\vspace{-10pt}
\end{center}
\caption{Additional results of FM-KT in the stronger strategies and stronger teacher setting.}
\label{tab:stronger_results}
\vspace{-10pt}
\end{table}

From Table~\ref{tab:stronger_results}, we can observe that FM-KT performs remarkably when both the teacher and the strategies are stronger. For instance, when $K$=8, the student's accuracy is 73.70\%, which is 0.82\% higher than the baseline \baseline{DIST}. This is a clear indication that FM-KT can be generalized to scenarios with strong strategies and a stronger teacher.

\section{Vision Transformer Comparison}
\label{sec:deit}
Transformer has been brilliantly turned into an infrastructure for numerous computer vision~\citep{SWIN-T} and natural language processing tasks~\citep{GPT3,GPT4}. In this section, we integrate FM-KT with the Vision Transformer (ViT) model DeiT~\citep{DeiT} and conduct comparative experiments between FM-KT and other prevalent knowledge distillation methods (\textit{e.g.} ViTKD) tailored for ViT. For outcomes derived from all the relevant algorithms in our experiments, the training hyperparameters follow the basic settings outlined in MMClassification~\citep{mmcls}. Additionally, we implement FM-KT considering DIST as the loss function and a 2-MLP as the meta-encoder. This implementation is simultaneously coupled with ViTKD~\citep{ViTKD} to facilitate combined distillation. Ultimately, he experimental results on ImageNet-1k are presented in Table~\ref{tab:deit_compare}.
\setlength{\tabcolsep}{4pt}
\begin{table}[!h]
\footnotesize
\renewcommand\arraystretch{0.9}
\begin{center}
\resizebox{1.0\textwidth}{!}{%
\begin{tabular}{lccccccc}
\hline
\noalign{\smallskip}
& Teacher (DeiT III-Small) & \baseline{Student (DeiT-Tiny)} & ViTKD & ViTKD+NKD & FM-KT ($K$=1) & FM-KT ($K$=2) & FM-KT ($K$=4) \\
\hline
\noalign{\smallskip}
Top-1 Acc. & 81.35\% & \baseline{74.42\%} & 76.06\% & 77.78\% & 77.84\% & 78.15\% & 78.15\% \\
\hline
\end{tabular}}
\vspace{-10pt}
\end{center}
\caption{Additional results of FM-KT in vision transformer comparison.}
\label{tab:deit_compare}
\vspace{-10pt}
\end{table}

From Table~\ref{tab:deit_compare}, we can illustrate that FM-KT is apparently effective on the ViT setting. Specifically, FM-KT consistently surpasses the \baseline{baseline} DeiT-Tiny, ViTKD, and ViTKD+KD in the final performance with all feasible $K$ during inference.

\section{Visualization of Sampling Trajectory}
\label{apd:visualization}
To elucidate the sampling mechanism of FM-KT, we utilize the student obtained by training the ResNet34-ResNet18 pair on ImageNet-1k and visualize the student output's sampling trajectory (\textit{i.e.}, $\{Z_{1-i/K}\}_{i=1}^{K}$, where $K$ is set as 8) during inference. Note that the loss function and meta-encoder are set as DIST and Swin-Transformer in training, respectively. Since general visualization methods are designed for feature maps in intermediate layers, it's challenging to demonstrate that better visualization directly correlates with improved image classification performance. Therefore, we employ the reliability histogram for the visualization of the sampling trajectory, thereby demonstrating that FM-KT, as an implicit ensemble method, indeed enhances the generalization ability of the student.

\begin{figure}[!h]
\begin{center}
\includegraphics[width=0.95\linewidth]{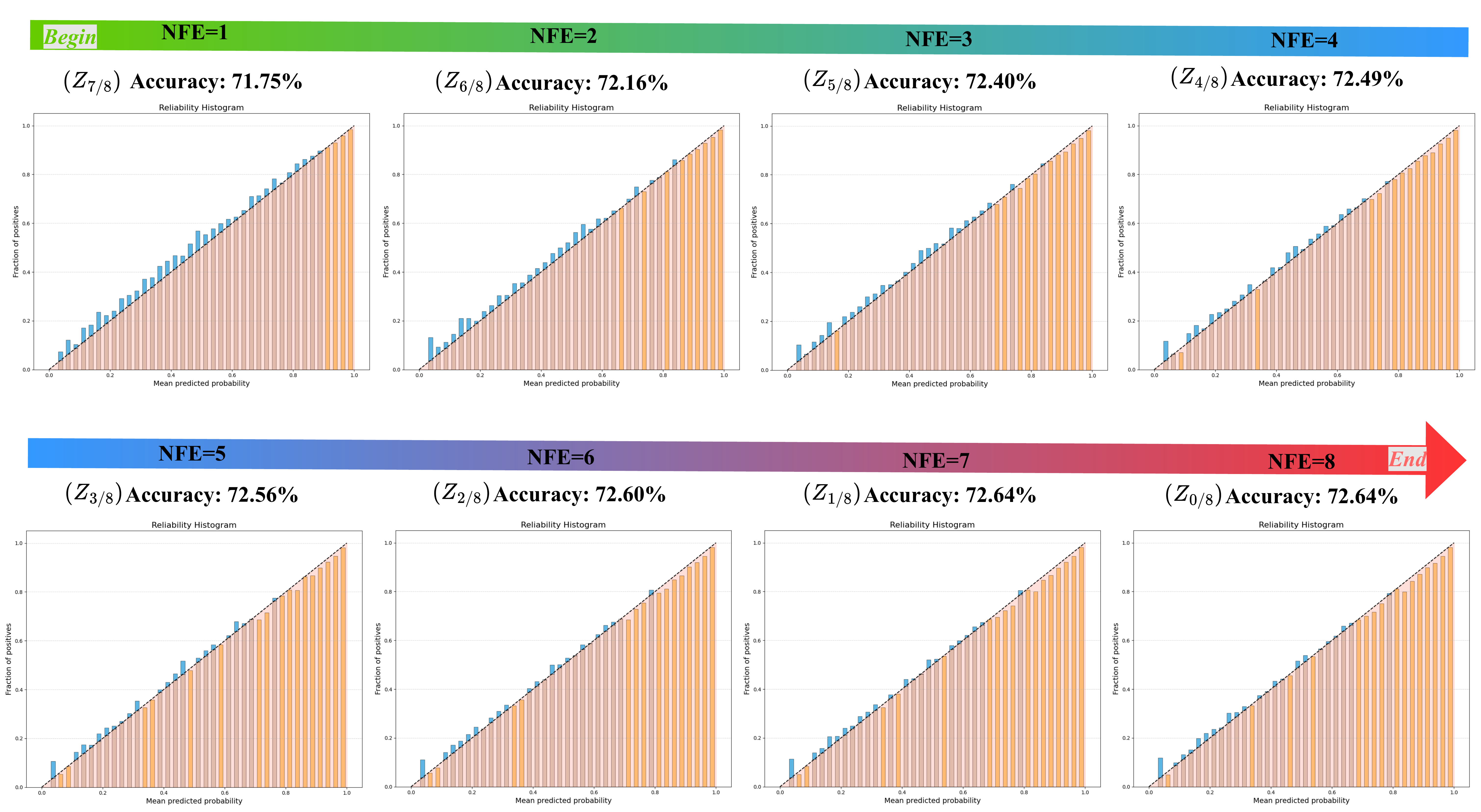}
\vspace{-10pt}
\end{center}
   \caption{The visualization of the sampling trajectory in the trained ResNet18.}
\label{fig:traj_visualization}
\vspace{-10pt}
\end{figure}

The reliability histogram typically plots predicted probabilities on the x-axis and the fraction of positives on the y-axis. Typically, the closer the predicted probability is to the fraction of positives, the better the student's prediction. Therefore, the closer the peak of the student's reliability histogram bin is to the diagonal, the stronger its generalization ability. Thus, it is clear from Fig.~\ref{fig:traj_visualization} that the reliability histogram is not well-presented at the beginning (\textit{i.e.} $Z_{7/8}$) of the sampling trajectory. As $i$ in $Z_{1-i/K}$ gradually decreases, the representation of its reliability histogram improves, which indicates that both the generalization ability and the reliability of the student are enhanced.

\section{Implementation Detail}
\label{apd:implementation_detail}

\subsection{Training Strategies}
\label{apd:implementation_detail:training_strategies}
We train on image classification {datasets} including CIFAR-100~\citep{CIFAR} and ImageNet-1k~\citep{ILSVRC15}, and object detection {dataset} including MS-COCO~\citep{COCO}. \textbf{For Offline KD}, the training strategy of image classification follows CRD~\citep{CRD} and DIST, while the training strategy of {object} detection follows PKD. Specifically, for CIFAR-100, the learning rate is 0.05 (when MobileNetV2 or ShuffleNetV1 as the student, the learning rate is 0.01), batch size is 64, total number of epochs is 240, and the learning rate is linearly reduced to 0.1 of its {previous} value at epochs 150, 180, and 210; for ImageNet-1k, the training learning rate is 0.1, batch size is 256, total number of epochs is 100, and the learning rate is linearly reduced to 0.1 of its {previous} value at epochs 30, 60, and 90; for MS-COCO, the training learning rate is 0.02, batch size is 16, total number of epochs is 24, and the learning rate is linearly reduced to 0.1 of its previous value at epochs 16 and 22. \textbf{For Online KD}, all hyperparameters settings follow AHBF-OKD~\citep{AHBF-OKD} and are unchanged. For conviction, we report the mean test accuracy with 3 runs for all experimental results.

\subsection{Loss Function and Meta-encoder}
\label{:apd:implementation_detail:loss_function}
The loss weights of FM-KT and its variant OFM-KT are not explicitly set, and their values will follow the loss weight settings of the metric-based distillation method introduced by themselves. For instance, if FM-KT applies DIST as its $L(\cdot,\cdot)$, the loss weights $\beta$ and $\gamma$ are both set to $2$ as mentioned in the original paper. For convenience of description, all forms ``FM-KT ($K$=number)'' or ``OFM-KT ($K$=number)'' refer to the corresponding algorithms that sampled ``number'' steps during inference. 

For all comparative experiments on CIFAR-100, FM-KT and OFM-KT use Swin-Transformer as the meta-encoder and DIST as the metric-based distillation method, except for VGG13-VGG8 and VGG13-MobileNetV2 pairs in the Offline KD scenario. VGG13-VGG8 and VGG13-MobileNetV2 pairs in the Offline KD scenario use Swin-Transformer as the meta-encoder and DKD as the metric-based distillation method.  For all comparative experiments on ImageNet-1k, in the Offline KD scenario, FM-KT uses MLP (\textit{i.e.} 2-MLP) as the meta-encoder and DIST as the metric-based distillation method; in the Online KD scenario, OFM-KT uses Swin-Transformer as the meta-encoder and DIST as the metric-based distillation method.

For FM-KT$^\Theta$, the loss function and meta-encoder are set to DKD and Swin-Transformer with all pairs on CIFAR-100; the loss function and meta-encoder are set to DIST and MLP with all pairs on ImageNet-1k; the balance weight $\alpha^{\Theta}$ is set as 1.0, 1.0 and 0.0 on all teacher-student pairs on CIFAR-100, ResNet34-ResNet18 pair on ImageNet-1k and ResNet50-MobileNetV2 pair on ImageNet-1k, respectively. 

For object detection, unless otherwise specified, FM-KT uses CNN as the meta-encoder and PKD as the metric-based distillation method. 

For the architecture of the meta-encoder, we adopt a {task-specific setup}. Swin-Transformer adopts one layer of \textbf{[Swin Attention-Linear-ReLU-Linear]} in the Offline KD scenario, and the number of heads is $4$. {In the Online KD scenario, if the student architecture is not ResNet18 then we add the same extra layer in the meta-encoder.} CNN uses one layer of \textbf{{[SiLU-Conv-GroupNorm-SiLU-Conv]}} in the image classification datasets and two layer of \textbf{{[Depthwise Conv-LayerNorm-Pointwise Conv-GeLU-Pointwise Conv]}} in the object detection dataset. In image classification, the kernel size of first convolutional layer is 3$\times$3, and the second layer is 1$\times$1. And in object detection, {the kernel size of the depthwise convolutional layer is 7$\times$7.} MLP adopts two layers of \textbf{[Linear-ReLU-Linear]} in the logit-based distillation scenario and one layer of \textbf{[Linear-ReLU-Linear]} in the feature-based distillation scenario. Besides, the shape transformation function $\mathcal{T}(\cdot)$ utilizes one layer of \textbf{{[Conv]}} or \textbf{[Identity Function]} (if no shape alignment is required) in the feature-based distillation scenario, and we use one layer of \textbf{[AdaptAvgpool(1)-Linear]} in the logit-based distillation scenario. Note that in the logit-based distillation scenario, FM-KT completes flow matching on the logit, so \textbf{[AdaptAvgpool(1)-Linear]} essentially represents the classification layer.

\section{Additional Training and Inference Computational Cost Discussion}
\label{apd:add_training_and_inference_cc_discussion}
\begin{figure}[!h]
\begin{center}
\includegraphics[width=0.95\linewidth]{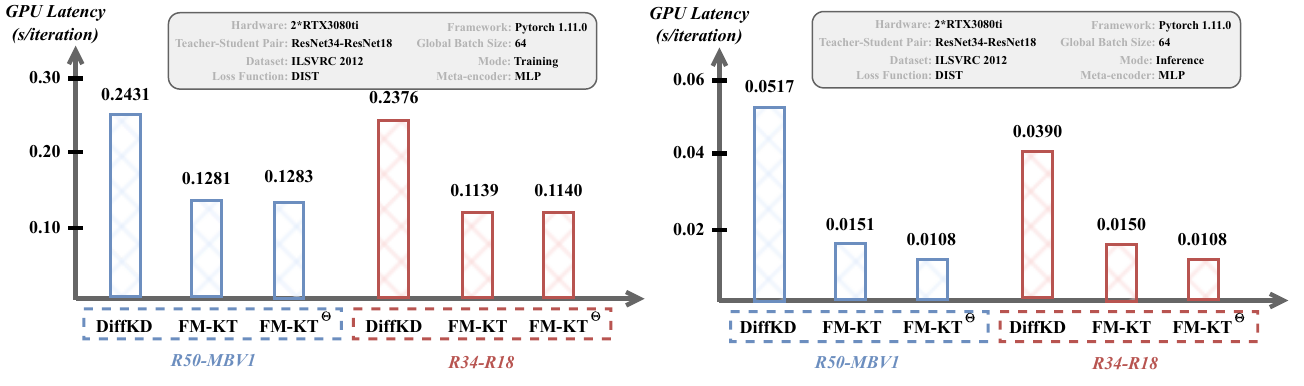}
\vspace{-10pt}
\end{center}
   \caption{Training and Inference computational cost of DiffKD, FM-KT and FM-KT$^{\Theta}$}
\label{fig:additional_training_and_inference_cost}
\vspace{-10pt}
\end{figure}
Our proposed FM-KT, similar to DiffKD, incurs an additional computational burden during inference. However, our variant, FM-KT$^{\Theta}$, differs in that it avoids this extra computational load during inference. This is achieved by transferring the knowledge in $Z_0$ (\textit{w.r.t.} $t$=0) in FM-KT to the vanilla classification head of the student. To provide a clear comparison of the computational costs of DiffKD, FM-KT, and FM-KT$^{\Theta}$, we have conducted relevant measurements. The results are presented in Fig.~\ref{fig:additional_training_and_inference_cost}. Notably, both FM-KT and FM-KT$^{\Theta}$ utilize logit-based distillation as the loss function (referred to as DIST) and employ a 2-layer MLP as the meta-encoder. Moreover, DiffKD adheres to the approach outlined in its original paper, employing both feature-based and logit-based distillation. The feature-based distillation in DiffKD, which relies on Bottleneck from ResNet, is implemented in the meta-encoder and is inserted into the backbone output feature before average pooling. Meanwhile, its logit-based distillation employs a 1-layer MLP as the meta-encoder and is inserted into the output logit of the classification head. As presented in Fig.~\ref{fig:additional_training_and_inference_cost}, the computational overhead of DiffKD, in both training and inference, is drastically higher than that of FM-KT and FM-KT$^{\Theta}$. Furthermore, FM-KT$^{\Theta}$ aligns with classical knowledge distillation algorithms in terms of computational cost during inference, offering additional savings in inference overhead compared to FM-KT.

\section{Best Meta-encoder Choice on ImageNet-1k}
\label{apd:best_meta_encoder}
As illustrated in Figure~\ref{table:ablation_loss_encoder}, FM-KT achieves the highest effectiveness and efficiency on ImageNet-1k when implemented with MLP. Accordingly, this section presents the optimal performance of FM-KT by using MLP for meta-encoder on ImageNet-1k and examines the impact of varying the number of MLP layers on its performance.
\setlength{\tabcolsep}{4pt}
\begin{table}[!h]
\footnotesize
\renewcommand\arraystretch{0.9}
\begin{center}
\begin{tabular}{l|cccc}
\hline
\noalign{\smallskip}
Method& FM-KT & FM-KT & FM-KT & DiffKD \\
Meta-encoder & 1-MLP & 2-MLP & 3-MLP & \makecell{2-BottleNeck+\\Conv+BN+MLP}\\
\hline
\noalign{\smallskip}
ResNet34-ResNet18 & 72.48 & 73.17 & \textbf{73.28} & 72.49 \\
ResNet50-MobileNetV1 & 73.74 & 74.22 & \textbf{74.28} & 73.78 \\
\hline
\end{tabular}
\vspace{-10pt}
\end{center}
\caption{The influence of the number of layers in the meta-encoder (\textit{i.e.} MLP) on student performance on ImageNet-1k. Note that all results from FM-KT are obtained when $K$=8.}
\label{tab:mlp_layer}
\vspace{-10pt}
\end{table}

The experimental results presented in Table~\ref{tab:mlp_layer} show that FM-KT outperforms DiffKD with 2-later MLP (\textit{i.e.} 2-MLP). Furthermore, as detailed in Appendix~\ref{apd:add_training_and_inference_cc_discussion}, the training and inference costs of FM-KT are nearly half those of DiffKD. This effectively demonstrates FM-KT's capability to not only outperform DiffKD but also achieve state-of-the-art performance. Meanwhile, the performance of FM-KT improves as the number of layers in the MLP increases.

\section{Architecture-Sensitive Experiments between FM-KT and DiffKD}
\label{apd:arch_sensitive}
\setlength{\tabcolsep}{4pt}
\begin{table}[!h]
\footnotesize
\renewcommand\arraystretch{0.9}
\begin{center}
\resizebox{0.95\textwidth}{!}{%
\begin{tabular}{cccc}
\hline
\noalign{\smallskip}
\makecell{DiffKD uses \\FM-KT's meta-encoder\\(\textit{i.e.} 2-MLP)} & \makecell{FM-KT uses \\DiffKD's meta-encoder\\(\textit{i.e.} 2-Bottleneck+Conv+BN+MLP)} & \makecell{DiffKD uses \\DiffKD's meta-encoder\\(\textit{i.e.} 2-Bottleneck+Conv+BN+MLP)} & \makecell{FM-KT uses \\FM-KT's meta-encoder\\(\textit{i.e.} 2-MLP)} \\
\hline
\noalign{\smallskip}
NAN & \textbf{74.26\%} & 73.78\% & 74.22\% \\
\hline
\end{tabular}}
\vspace{-10pt}
\end{center}
\caption{Comparison experimental result between FM-KT and DiffKD with ResNet50-MobileNetV1 pair on ImageNet-1k. Note that the results from FM-KT are obtained when $K$=8.}
\label{tab:comparison_fm_kd_diffkd}
\vspace{-10pt}
\end{table}
In order to know the sensitivity of FM-KT and DiffKD to architecture and for further fair comparisons, we perform DiffKD to use 2-MLP from FM-KT as its meta-encoder, and FM-KT to use 2-Bottleneck+Conv+BN+MLP from DiffKD as its meta-encoder. The experiments were then conducted on ImageNet-1k using ResNet50-MobileNetV1 pair. Unfortunately, when employing the logit-based distillation approach of DiffKD (following its official code and implementation~\citep{DiffKD}), its loss became NAN at epoch 1. However, in Table~\ref{tab:comparison_fm_kd_diffkd}, we discover that the result (with $K$=8) of FM-KT using 2-Bottleneck+Conv+BN+MLP from DiffKD as its meta-encoder, which significantly outperformed DiffKD.

\section{Unify VP SDE, VE SDE and Rectified flow in FM-KT}
\label{apd:unified}
Vanilla diffusion processes such as VP SDE~\citep{sde} and VE SDE~\citep{sde} and vanilla continuous probability flows such as Rectified flow~\citep{iclr22_rect} can be transformed into our proposed FM-KT\footnote{Both VP SDE and VP SDE can be transformed into ODE form, referring to deterministic forward and backward processes.}. Here, we give the derivation of FM-KT. All noise schedule can be written in the following form:
\begin{equation}
\fontsize{8pt}{11pt}\selectfont
\begin{aligned}
& Z_t = \alpha_t X^S + \sigma_t X^T,\ s.t.\ \lim_{t\rightarrow 0}\alpha_t = 0, \lim_{t\rightarrow 0}\sigma_t = 1, \lim_{t\rightarrow 1}\sigma_t = 0.\\
\end{aligned}
\label{eq:general_fm_kd_1}
\end{equation}
The training paradigm of them can be denoted as
\begin{equation}
\fontsize{8pt}{11pt}\selectfont
\begin{aligned}
& \operatorname*{arg\,min}_{v_\theta}\mathbb{E}_{(Z_1,Z_0,t)}||g_{v_\theta}(Z_t,t) - \nabla_t Z_t||_2^2\\
& = \operatorname*{arg\,min}_{v_\theta}\mathbb{E}_{(Z_1,Z_0,t)}||g_{v_\theta}(Z_t,t) - (\nabla_t\alpha_t X^S + \nabla_t\sigma_t X^T)||_2^2.\\
\end{aligned}
\label{eq:general_fm_kd_2}
\end{equation}
\paragraph{VP ODE:} (1) $\alpha_t = \textrm{exp}(-\frac{1}{4}a(1-t)^2-\frac{1}{2}b(1-t))$; (2) $\sigma_t = \sqrt{1-\alpha_t^2},\ s..t.\quad a=19.9, b=0.1$.
\paragraph{VE ODE:} (1) $\alpha_t =  a(\frac{b}{a})^{t}$; (2) $\sigma_t = 1,\ s.t.\quad a=0.02,b=100$.
\paragraph{Rectified flow:} (1) $\alpha_t = t$; (2) $\sigma_t = 1-t$.

Substituting $\alpha_t$ and $\sigma_t$ yields:

\textbf{VP ODE:}
\begin{equation}
\fontsize{8pt}{11pt}\selectfont
\begin{aligned}
& \operatorname*{arg\,min}_{v_\theta}\mathbb{E}_{(Z_1,Z_0,t)}||g_{v_\theta}(Z_t,t) - ((\frac{1}{2}a(1-t)+\frac{1}{2}b)\alpha_t X^S - \frac{\alpha_t}{\sqrt{1-\alpha_t^2}}\alpha_t(\frac{1}{2}a(1-t)+\frac{1}{2}b) X^T )||_2^2. \\
\end{aligned}
\label{eq:general_fm_kd_3}
\end{equation}

\textbf{VE ODE:}
\begin{equation}
\fontsize{8pt}{11pt}\selectfont
\begin{aligned}
& \operatorname*{arg\,min}_{v_\theta}\mathbb{E}_{(Z_1,Z_0,t)}||g_{v_\theta}(Z_t,t) - (\alpha_t[\log(b)-\log(a)] X^S )||_2^2. \\
\end{aligned}
\label{eq:general_fm_kd_4}
\end{equation}

\textbf{Rectified flow:}
\begin{equation}
\fontsize{8pt}{11pt}\selectfont
\begin{aligned}
& \operatorname*{arg\,min}_{v_\theta}\mathbb{E}_{(Z_1,Z_0,t)}||g_{v_\theta}(Z_t,t) - (X^S - X^T)||_2^2. \\
\end{aligned}
\label{eq:general_fm_kd_5}
\end{equation}

All forms can be transformed into serial training forms by Theorem~\ref{the:training_paradiam} in our paper\footnote{For convenience, we ignore time steps here. It is worth noting that, due to the adaptability of step size in the Euler method, introducing this hyperparameter is entirely feasible.}:
\begin{equation}
\fontsize{8pt}{11pt}\selectfont
\begin{aligned}
 &\mathcal{L}_\textrm{FM-KT++} = \mathbb{E}_{(X^S,X^T,Y)} \frac{1}{N}\sum_{i=0}^{N-1} L(\mathcal{T}((\nabla_t\alpha_t Z_1 - g_{v_\theta}(Z_{1-i/N},1-i/N))/-\nabla_t\sigma_t), X^T) \\
 & + \underbrace{L(\mathcal{T}((\nabla_t\alpha_tZ_1 - g_{v_\theta}(Z_{1-i/N},1-i/N))/-\nabla_t\sigma_t), Y)}_{\textrm{match the ground truth label (optional)}},\\
& \textrm{the sampling process:}\quad Z_{1-i/N} = Z_{1-(i-1)/N} - g_{v_\theta}(Z_{1-(i-1)/N},1-(i-1)/N)/N,\quad s.t.\quad i\geq1,
\end{aligned}
\label{eq:general_fm_kd_6}
\end{equation}
where $Z_1=\alpha_1X_S$. Thus, the key to achieving knowledge transfer in knowledge distillation is not the difference between noise schedules, but the form of deterministic sampling in both the forward and backward processes and the serial training paradigm given in Theorem~\ref{the:training_paradiam} of our paper.

\begin{figure}[!h]
\begin{center}
\includegraphics[width=1.0\linewidth]{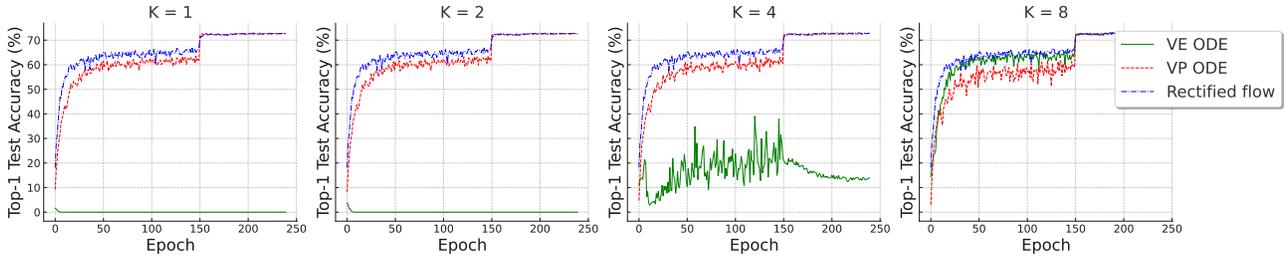}
\end{center}
\vspace{-0.2in}
   \caption{Trajectories of Top-1 test accuracy with WRN-40-2-WRN-16-2 pair on CIFAR-100 for various noise schedules: VP ODE, VE ODE, and Rectified flow.}
\label{fig:noise_schedule}
\vspace{-10pt}
\end{figure}

\noindent{\bf Evaluation.} In the practical implementation, since $\lim_{t\rightarrow 1}\nabla_t\alpha_t = +\infty$ in VP ODE, and considering that $\nabla_t\alpha_t$ and $\nabla_t\sigma_t$ show large variations at different $t$ in both VP ODE and VE ODE, both VP ODE and VE ODE are expressed in the forms of differentiations $\frac{\alpha_t - \alpha_{t - \Delta t}}{t - \Delta t}$ and $\frac{\sigma_t - \sigma_{t - \Delta t}}{t - \Delta t}$. Since $\nabla_t \sigma_t \equiv 0$ in VE ODE cannot be divided, we modified $\sigma_t$ from $\sigma(t)=1$ to $\sigma(t)=1-0.1t$. In addition, our experiments revealed instability in the flow loss of VE ODE and VP ODE training, necessitating the use of the learning rate warm-up technique (extending up to 20 epochs) for effective training. And $b$ in VE ODE is extra reduced to $10$. The test accuracy per epoch for VP ODE, VE ODE and Rectified flow (\textit{i.e.}, the default form used in our paper) is illustrated in Fig.~\ref{fig:noise_schedule}, which is the same as Fig.~\ref{fig:noise_schedule_main_paper} in our main paper.

The experimental results in Fig.~\ref{fig:noise_schedule} are obtained on CIFAR-100 with WRN-40-2-WRN-16-2 pair. VP ODE, VE ODE, and Rectified flow all utilize 2-MLP as the meta-encoder, and DIST as the loss function (modifying the hyperparameter temperature to 1 for stable training). It can be observed that the training paradigm proposed in Eq.~\ref{eq:general_fm_kd_6} is capable of effectively training all noise schedules. In particular, Rectified flow is comparatively more stable and efficient than VP ODE and VE ODE.

\noindent{\bf Discussion.} Through the above analysis, coupled with our new insights, the reasons for adopting Rectified flow in our work are as follows: 1) Rectified flow is more effective and stable compared to VP ODE and VE ODE; 2) Rectified flow is simple for implementation and understanding; 3) in the derivation of approximating ensembles (\textit{i.e.} Proposition~\ref{pps:link_ensemble_with_flow}), Rectified flow can be proved that the truncation error at each time step has the same impact on the ultimate error (\textit{i.e.}, equal weight); 4) Rectified flow enhances the student performance by its accelerated sampling property when NFE is very small. Specifically, Rectified flow has the ability to minimize the hessian matrix~\citep{icml23_curvature} with respect to $Z_t$, which enables the estimation $g_{v_\theta}(Z_t,t)$ of $dZ_t$ to also accurately estimate $\{dZ_{t-\Delta_t},dZ_{t-2\Delta_t},\cdots,dZ_{s}\}$, where $t$ and $s$ refer to the source and target time points, respectively, ultimately reducing the truncation error of $Z_t + \int_t^{s}g_{v_\theta}(Z_\tau,\tau)d\tau$. This point is demonstrated by the experiments in relevant papers~\citep{iclr22_rect,icml23_curvature}.

\section{Limitation}
\label{apd:limitation}
FM-KT demonstrates improved generalization capabilities relative to conventional knowledge distillation methods, yet it incurs a higher computational burden during inference. Moreover, FM-KT's effectiveness in object detection is not as pronounced as in image classification. This discrepancy stems from the fact that, in image classification, flow matching with the teacher at the logit level often yields performance akin to the teacher's. In contrast, in object detection, flow matching with the teacher at the FPN (Feature Pyramid Network) level does not directly translate to enhanced performance in the ultimate metric, mAP.

\end{document}